%% file: main.tex
\documentclass[11pt]{article}

\usepackage[final]{acl}

\usepackage{times}
\usepackage{latexsym}
\usepackage[T1]{fontenc}
\usepackage[utf8]{inputenc}
\usepackage{microtype}

\usepackage{amsmath}
\usepackage{amssymb}
\usepackage{mathtools}
\usepackage{amsthm}

\usepackage{graphicx}
\usepackage{subcaption}
\usepackage{booktabs}
\usepackage{float}
\IfFileExists{placeins.sty}{\usepackage{placeins}}{}
\usepackage{longtable}
\usepackage{tikz}
\usetikzlibrary{arrows.meta, positioning, calc, fit, backgrounds, shadows}
\IfFileExists{pdflscape.sty}{\usepackage{pdflscape}}{}
\usepackage{array}

\usepackage[capitalize,noabbrev]{cleveref}

\usepackage{algorithm}
\IfFileExists{algpseudocode.sty}{\usepackage{algpseudocode}}{}

\newcommand{\showcomments}{0} %
\newcommand{\authcomment}[3]{%
\ifnum\showcomments=1
  {\textcolor{#2}{\textbf{[#1: #3]}}}%
\fi}

\usepackage{inconsolata}
\makeatletter
\input{t1zi4.fd}
\makeatother
\DeclareFontShape{T1}{zi4}{m}{it}{<->ssub * zi4/m/n}{}
\usepackage{listings}
\usepackage{fancyvrb}
\usepackage{pifont}
\newcommand{\cmark}{\ding{51}}
\newcommand{\xmark}{\ding{55}}
\DefineVerbatimEnvironment{PromptVerb}{Verbatim}{breaklines,fontsize=\small}

\usepackage{enumitem}

\title{A Dataset for Modeling Iterative Problem-Solving}
\author{%
  Fagun Patel\textsuperscript{1*} \quad
  Sang T. Truong\textsuperscript{1*} \quad
  Duc Q. Nguyen\textsuperscript{2*} \quad
  Kazunori Fukuhara\textsuperscript{1} \\
  \bfseries Benjamin W. Domingue\textsuperscript{1\textdagger} \quad
  Sanmi Koyejo\textsuperscript{1\textdagger} \quad
  Nick Haber\textsuperscript{1\textdagger} \\[2pt]
  \textsuperscript{1}Stanford University \quad
  \textsuperscript{2}National University of Singapore \\[2pt]
  {\small \textsuperscript{*}Co-first authors}
}

\begin{document}
\maketitle

\input{sections/0abstract}

\input{sections/1intro}

\input{sections/2relatedworks}

\input{sections/4model}

\input{sections/3data}
\input{sections/6Result}

\input{sections/7Conclusion}

{\sloppy
\bibliography{references}
}

\clearpage
\appendix
\input{sections/8.10A10.DatasetDetails}

\input{sections/8.7A7.StudentTrajectories}

\input{sections/8.11A11.TrainingDetails}

\input{sections/8.9A9.LLMPredictorDetails}

\input{sections/8.1A1.UseofGenerativeAITools}

\end{document}

%% file: sections/0abstract.tex
\begin{abstract}
    Solving problems through repeated attempts is a sequential modeling task: at each step, the solver receives feedback and decides how to revise their solutions. Predicting whether performance improves, plateaus, or regresses across attempts is central to understanding any iterative problem-solving process in both human learners and autonomous agents. Beyond outcomes, modeling what errors persist and how strategies shift across attempts provides deeper insight into the mechanics of sequential learning. Studying these dynamics requires observing many solvers as they attempt, receive feedback, and revise. Programming courses with automated grading provide this setting, as students iteratively submit code to test suites and receive feedback on every attempt. We therefore curate CodeInsight, a large-scale dataset of over 3 million submissions from 3,286 undergraduates across 2 introductory C++ courses in 2 academic years, with test-case-level outcomes, timestamps, and source code. On this dataset, we build a benchmark that evaluates models spanning parametric, sequential, and generative traditions under a shared calibration-and-scoring protocol, including a Recurrent State Space Model (RSSM) adapted to track solver characteristics through discrete latent variables and an LLM-based predictor that generates explicit solutions. The adapted RSSM achieves the strongest predictive accuracy on three of the four courses. The LLM predictor is less accurate but produces full submissions at each attempt, enabling direct analysis of failure modes. We find that the model's coding proficiency is inversely related to predictive performance in this setting, with the LLM better understood as a generative solver conditioned on context rather than a faithful predictor of solver behavior. We publicly release our code and the dataset on request to facilitate future research.\footnote{Code: \href{https://github.com/aims-foundations/dynamic-irt}{\nolinkurl{github.com/aims-foundations/dynamic-irt}}; Dataset: \href{https://huggingface.co/datasets/CodeInsightTeam/code_insights_csv}{\nolinkurl{huggingface.co/datasets/CodeInsightTeam/code_insights_csv}}.} %
\end{abstract}

%% file: sections/1intro.tex
\section{Introduction}
Problem-solving is inherently iterative: a solver receives feedback, revises their solutions, and tries again. The resulting trajectory of attempts encodes how their solutions evolve. Students submitting code against a test suite or LLMs solving a coding task both produce sequential records of iterative refinement~\citep{Fahid2021, Rivers2016LCA}. These trajectories are rich but noisy: a single-character change can flip a submission from incorrect to correct, and solvers may improve steadily, plateau for long stretches, or regress after apparent progress on a given task. Predicting these dynamics is a sequential modeling problem with practical applications from evaluating autonomous agents to adaptive instruction~\citep{piech2015deepknowledgetracing}. Beyond predicting outcomes, modeling which errors persist and how strategies shift across attempts can inform personalized feedback, diagnose systematic failure modes, and support use cases such as routing model decisions on LLMs~\citep{RouteLLM2025, FrugalGPT2023}.

Several families of models have been applied to predicting problem-solving trajectories. Parametric models such as Item Response Theory (IRT)~\cite{embretson2000item} estimate fixed solver and problem parameters to predict binary outcomes, but do not typically model how solvers' skill levels evolve over attempts. Sequential prediction models track how solver performance changes but remain limited to binary correctness; Bayesian Knowledge Tracing (BKT)~\cite{CorbettAndAnderson1995}, a hidden Markov model, and Deep Knowledge Tracing (DKT)~\cite{piech2015deepknowledgetracing}, a type of Recurrent Neural Network, are the primary examples in this domain. More recently, LLMs have been used to emulate problem-solving behavior~\cite{Markel2023}, generating solutions~\cite{scarlatos_smart_2025} and operating over richer representations that can capture failure modes directly from context. These approaches have developed independently, and no existing study compares them for next-step prediction under a shared evaluation framework.

Programming courses provide rich temporal data for such a comparison: students submit code repeatedly against automated test suites, receive immediate feedback, and generate longitudinal trajectories that span the full arc of iterative problem-solving~\citep{Messer2024, Blikstein2014}. We curate CodeInsight, a large-scale dataset of 3,286 undergraduates learning C++ across 394 problems at a Vietnamese university (2022--2024), comprising over 3 million submissions with test-case-level correctness, timestamps, and source code. On this dataset we build a benchmark that evaluates models spanning parametric, sequential, and generative traditions under a shared calibration-and-scoring protocol. Among these, we adapt a Recurrent State Space Model (RSSM) with discrete latent variables that tracks solver characteristics over time, and an LLM-based predictor that generates the solutions based on students' profiles. We find that RSSM achieves stronger predictive accuracy on three of the four courses by training directly on response patterns, while the LLM predictor falls below all trained models. Because the LLM generates complete solution attempts, we can diagnose this gap directly: its predictive signal is driven more by problem difficulty than by solver-specific behavior, and models with stronger coding ability predict worse. Our contributions are as follows:
\begin{itemize}[left=0pt, itemsep=1pt, topsep=0pt]
    \item \textbf{Dataset.} We curate CodeInsight, a large-scale dataset of 3,286 undergraduates learning C++ across 394 problems at a Vietnamese university (2022--2024), comprising over 3 million code submissions with test-case-level correctness, timestamps, and source code, released under a departmental license upon request.
    \item \textbf{Benchmarking Predictive Models.} We introduce a benchmark for next-attempt prediction on sequential data that evaluates models spanning parametric, sequential, and generative traditions under a shared calibration-and-scoring protocol; to our knowledge, this is the first study to compare these modeling traditions under one evaluation framework. The RSSM achieves the highest AUC on three of the four courses, while the LLM predictor falls below the parametric models and the RSSM at nearly every attempt, with a predictive signal driven more by problem difficulty than solver-specific behavior.
\end{itemize}

%% file: sections/2relatedworks.tex
\section{Related Works}
\paragraph{Programming trajectory datasets}
Several datasets capture student interactions in programming courses, though they differ in granularity. CodeWorkout~\cite{DataShop2021_CodeWorkout} provides ${\sim}$130K Java submissions from ${\sim}$820 students with source code, timestamps, and aggregate test-pass percentages, but not per-test-case outcomes. The Blackbox dataset~\cite{Brown2018FiveYears} logged over 300M compilations and source code edits from 2.5M BlueJ users worldwide, but lacks assignment structure or correctness labels; the project has since been discontinued, and all data access will be revoked by August 2026. ACcoding~\cite{chen2024accoding} provides 4M submissions across multiple languages from 27K students, but records a single overall judge verdict per submission. CodeInsight differs from these in providing per-test-case pass/fail labels across 3,888 individual test cases, full source code, and timestamps for over 3M submissions, enabling fine-grained analysis of how solvers fail on specific subtasks across attempts.

\paragraph{Modeling and Predicting Iterative Problem-Solving}
Modeling how problems get solved through iterative attempts has drawn on several traditions. In repetitive-practice settings, \citet{reddy2016unbounded} propose a modified exponential forgetting curve and explore how task difficulty and the number of repetitions influence learning efficiency, validating these concepts through large-scale log data from a flashcard system. \citet{mohr2018deterministic} show how performance dynamics are shaped by repetitive practice and feedback, illustrating how solvers gradually adjust their strategies based on the outcomes of prior attempts. Within item response theory, sequential IRT models describe repeated attempts at the same item~\citep{culpepper2014sequential}, with later extensions to cognitive diagnosis and adaptive testing under repeated attempts~\citep{hung2019sequential, lu2026adaptive}; our CIRT baseline follows this tradition by adding an attempt-indexed difficulty slope to IRT. In programming specifically, recent work extends knowledge tracing models, which predict a solver's next outcome from their interaction history~\citep{shen2024ktsurvey}, with richer code representations. Code-DKT~\citep{shi2022code_dkt} incorporates AST-based code representations via code2vec, and Temporal-ASTNN~\citep{mao2021temporalastnn} combines AST-based neural networks with LSTMs to model temporal progression. TIKTOC~\citep{duan2025} predicts per-test-case outcomes rather than aggregate correctness. Together with OKT~\citep{liu-etal-2022-open}, they are able to generate student code via a language model conditioned on a latent state. However, no existing study compares code-aware sequential models, parametric models, and generative approaches under a shared evaluation framework, which our work addresses.

\begin{figure*}[t]
    \centering
    \includegraphics[width=\textwidth]{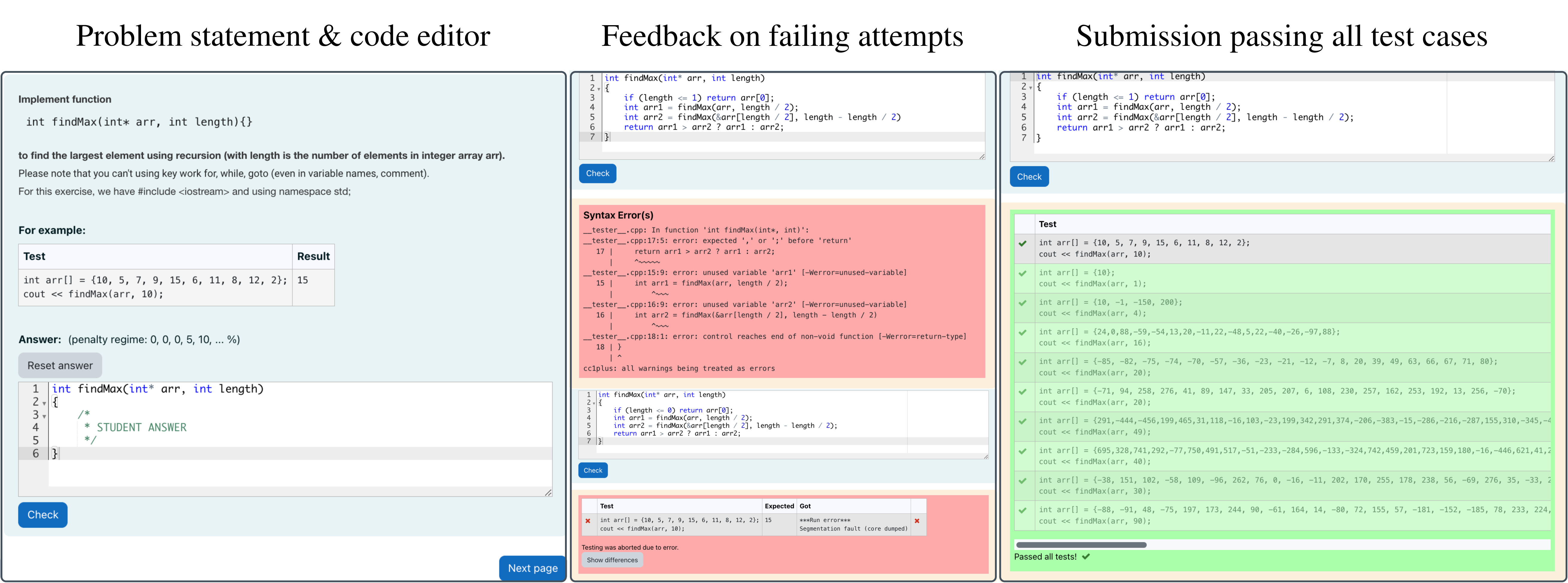}
    \caption{The exercise platform interface based on Moodle. From left to right: the problem statement with function signature, description, example test case, and the code editor where the solution attempt is written; automated feedback on failing attempts, with compiler errors (red) and failed test cases showing expected versus actual output; and a revised submission where all test cases pass (green).}
    \label{fig:data-ui}
\end{figure*}

\paragraph{LLMs as simulated problem-solvers}
LLMs have been increasingly used to simulate learner behavior. \citet{aher2023usinglargelanguagemodels} introduced Turing Experiments to evaluate how well LLMs replicate human behavioral patterns. In programming education, \citet{wu2025embracingimperfectionsimulatingstudents} proposed constructing cognitive prototypes via knowledge graphs to constrain LLM output toward realistic errors; \citet{Markel2023} developed GPTeach, a system that uses GPT-simulated students for training teaching assistants, and \citet{miroyan_parastudent_2025} generated realistic student code by fine-tuning on timestamped submission histories. \citet{ross2025learningmakemistakesmodeling} proposed a method for improving student error simulation through cycle-consistent reasoning between incorrect answers and latent misconceptions. On the prediction side, \citet{neshaei2024towards} found that fine-tuned LLMs match but do not surpass sequential models on next-step prediction. Whether LLMs capture solver-specific failure modes or primarily reflect their own coding ability remains an open question that we address directly in this work.

%% file: sections/4model.tex
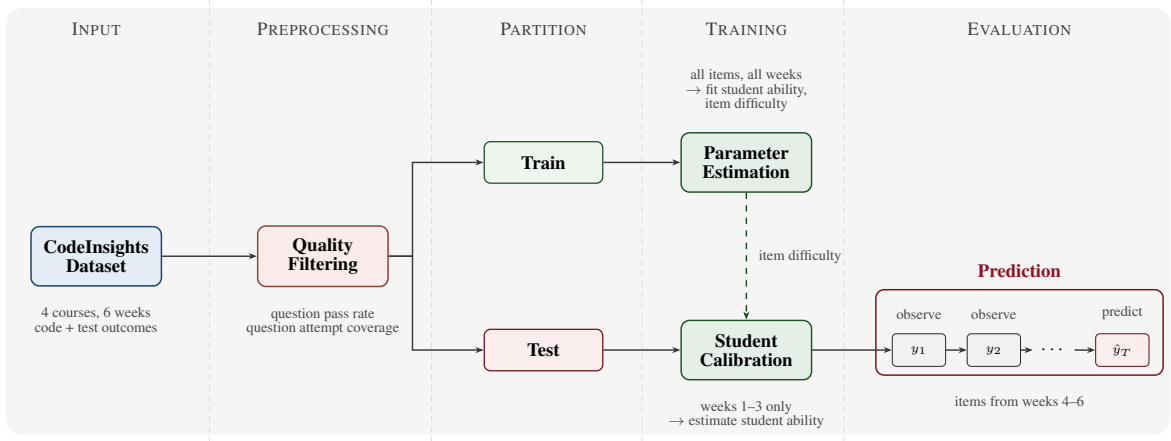
\begin{figure*}[t]
  \centering
  \resizebox{\textwidth}{!}{\input{figures/eval_pipeline}}
  \caption{Shared evaluation pipeline. After quality filtering, students are partitioned into train and test sets. Train students to estimate item difficulty parameters ($b_j$) across all weeks. These frozen item parameters are transferred to the test set, where student ability ($\theta_{\mathrm{test}}$) is calibrated in the early weeks. The calibrated model then predicts test student performance on held-out weeks. All models follow this pipeline.}
  \label{fig:eval_pipeline}
\end{figure*}

\section{Dynamic Measurement Models}
\label{sec:models}
We evaluate whether student performance on future problems can be predicted from prior submission history. We compare parametric and sequential baselines with two adapted approaches: an RSSM and an LLM-based predictor.

\paragraph{Baseline models}
We compare against six baselines spanning two families. The parametric models are IRT, a one-parameter logistic model with a static per-solver ability and a per-test-case difficulty~\cite{embretson2000item} (Appendix~\ref{sec:apd_irt}), and CIRT, a temporal extension of IRT that adds an attempt-indexed difficulty slope (Appendix~\ref{sec:apd_cirt}). The sequential models are BKT, a hidden Markov model with a binary mastery state per test case~\cite{CorbettAndAnderson1995} (Appendix~\ref{sec:apd_bkt}); DKT, a recurrent neural network over the interaction sequence~\cite{piech2015deepknowledgetracing} (Appendix~\ref{sec:apd_dkt}); Code-DKT, a code-aware DKT variant that augments each interaction with an embedding of the submitted code~\cite{shi2022code_dkt} (Appendix~\ref{sec:apd_codedkt}); and TIKTOC, a multi-task model that couples an LSTM knowledge state with a LoRA-finetuned language model to jointly predict per-test-case outcomes and generate the student's next submission~\cite{duan2025} (Appendix~\ref{sec:apd_tiktoc}). Training procedures and loss curves for all models are provided in Appendix~\ref{sec:training_details}.

\paragraph{Recurrent State Space Model (RSSM)}
\label{sec:rssm}
We introduce an RSSM that captures solver characteristics through discrete latent variables $z_t$ at each timestep. At each timestep $t$, both the solver's submitted code and the problem description are embedded using a pretrained LLM (Qwen3-Embedding~\cite{qwen3embedding}) and projected through separate learned linear layers into a response encoding $e_t$ and a problem encoding $q_t$. A GRU backbone~\cite{cho-etal-2014-learning} updates a hidden state $h_t = f_\phi(h_{t-1}, z_{t-1}, q_t)$ that captures the solver's cumulative history. A representation model (posterior) $q_\phi(z_t \mid h_t, e_t)$ infers discrete latent variables from the hidden state and the current response, while a transition predictor (prior) $p_\phi(z_t \mid h_t)$ predicts them from history alone. This posterior/prior separation follows the DreamerV2 framework~\cite{hafner2021mastering}: during training, the posterior grounds $z_t$ in the observed response, while at inference, the prior predicts solver characteristics before observing the next submission. A scoring head predicts per-test-case binary pass/fail outcomes from $q_t$, $h_t$, and $z_t$, combined with a learned per-item difficulty bias. The model does not generate code; it predicts test-case-level correctness. An embedding predictor provides an auxiliary training signal by predicting the next response embedding from $z_t$. Full architecture details and the loss function are provided in Appendix~\ref{sec:rssm_training}.

\paragraph{LLM-as-Predictor}
We investigate whether a pre-trained LLM, conditioned on a student's behavioral profile and submission history, can predict performance on future problems for held-out students. The LLM predictor follows the same train/test split and temporal protocol as the baselines. During testing, for each test-split student, the model receives a structured behavioral persona capturing submission pacing, precheck usage, topic-level pass rates, and code complexity. It also receives problem-level statistics derived from students in the training set (pass rate, typical attempt count) and retrieved examples from the student's own calibration history on similar problems, condensed into behavioral summaries by a smaller LLM. Appendix~\ref{sec:llm_predictor_appendix} describes how each of these inputs is computed and shows the full prompt template.

Predictions of the LLMs are temporally grounded. Particularly, at attempt $t$, the models receive the student's full trajectory through attempt $t{-}1$, including pass patterns and code summaries, then generate code predicting what the student would submit next. No fine-tuning is performed; the LLM parameters are frozen throughout. This asymmetry with the trained baselines is deliberate: the LLM serves as a zero-training reference point that tests whether an off-the-shelf model can predict solver behavior from context alone, and our claims about the LLM are scoped to this frozen setting. Prior work finds that fine-tuned LLMs match but do not surpass sequential models on next-step prediction~\citep{neshaei2024towards}. Full details of the retrieval mechanism, prompt construction, and simulation protocol are provided in Appendix~\ref{sec:llm_predictor_appendix}.

%% file: figures/eval_pipeline.tex
\definecolor{cData}{HTML}{2166AC}      %
\definecolor{cFilter}{HTML}{D6604D}    %
\definecolor{cSplit}{HTML}{762A83}     %
\definecolor{cCalib}{HTML}{2D7D3A}    %
\definecolor{cPred}{HTML}{B2182B}     %
\definecolor{cGray}{HTML}{525252}     %
\definecolor{cBG}{HTML}{F5F5F5}       %

\begin{tikzpicture}[
  >={Stealth[length=4pt, width=3pt]},
  every node/.style={font=\rmfamily},
  stage/.style={
    draw=#1!65!black, fill=#1!12, rounded corners=4pt,
    line width=0.7pt, minimum width=2.2cm, minimum height=1.0cm,
    align=center, font=\small\bfseries, inner sep=6pt,
    drop shadow={shadow xshift=0.4pt, shadow yshift=-0.4pt, fill=black!12}
  },
  branch/.style={
    draw=#1!60!black, fill=#1!8, rounded corners=3pt,
    line width=0.6pt, minimum width=2.0cm, minimum height=0.7cm,
    align=center, font=\footnotesize\bfseries, inner sep=4pt
  },
  annot/.style={
    font=\scriptsize, text=cGray!90!black, align=center
  },
  stlabel/.style={
    font=\footnotesize\scshape, text=cGray!80!black
  },
  arr/.style={->, line width=0.7pt, color=cGray!70!black},
  arrdash/.style={->, line width=0.7pt, color=cCalib!60!black, dashed},
]

\node[stage=cData] (dataset) {CodeInsights\\[-1pt]Dataset};
\node[annot, below=6pt of dataset] (ann-data) {4 courses, 6 weeks\\[-1pt]code + test outcomes};

\node[stage=cFilter, right=1.6cm of dataset] (filter) {Quality\\[-1pt]Filtering};
\node[annot, below=6pt of filter] (ann-filter) {question pass rate\\[-1pt]question attempt coverage};

\node[branch=cCalib, above right=0.7cm and 1.6cm of filter] (train) {Train};
\node[branch=cPred,  below right=0.7cm and 1.6cm of filter] (test)  {Test};

\node[stage=cCalib, right=1.3cm of train] (calib_train) {Parameter\\[-1pt]Estimation};
\node[annot, above=6pt of calib_train] (ann-ct) {all items, all weeks\\[-1pt]$\rightarrow$ fit student ability,\\[-1pt]item difficulty};

\node[stage=cCalib, right=1.3cm of test] (calib_test) {Student\\[-1pt]Calibration};
\node[annot, below=6pt of calib_test] (ann-ctest) {weeks 1--3 only\\[-1pt]$\rightarrow$ estimate student ability};

\coordinate (pred_anchor) at ([xshift=1.8cm]calib_test.east);

\node[draw=cGray!50!black, fill=cGray!8, rounded corners=2pt,
      minimum width=0.9cm, minimum height=0.55cm,
      font=\tiny\bfseries, align=center] (obs1) at (pred_anchor) {$y_1$};
\node[draw=cGray!50!black, fill=cGray!8, rounded corners=2pt,
      minimum width=0.9cm, minimum height=0.55cm,
      font=\tiny\bfseries, align=center,
      right=0.35cm of obs1] (obs2) {$y_2$};
\node[font=\small\bfseries, right=0.2cm of obs2] (dots) {$\cdots$};
\node[draw=cPred!50!black, fill=cPred!8, rounded corners=2pt,
      minimum width=0.9cm, minimum height=0.55cm,
      font=\tiny\bfseries, align=center,
      right=0.35cm of dots] (pred1) {$\hat{y}_T$};

\node[annot, above=3pt of obs1] (lbl1) {observe};
\node[annot, above=3pt of obs2] (lbl2) {observe};
\node[annot, above=3pt of pred1] (lbl3) {predict};

\draw[arr] (obs1) -- (obs2);
\draw[arr] (obs2) -- (dots);
\draw[arr] (dots) -- (pred1);

\node[draw=cPred!65!black, rounded corners=4pt, line width=0.7pt,
      fill=none, inner xsep=6pt, inner ysep=4pt,
      fit=(obs1)(obs2)(dots)(pred1)(lbl1)(lbl2)(lbl3)] (predict) {};

\node[font=\small\bfseries, text=cPred!80!black, above=2pt of predict] (pred_title) {Prediction};

\node[annot, below=6pt of predict] (ann-pred) {items from weeks 4--6};

\draw[arr] (dataset) -- (filter);
\draw[arr] (filter.east) -- ++(0.4,0) |- (train.west);
\draw[arr] (filter.east) -- ++(0.4,0) |- (test.west);
\draw[arr] (train) -- (calib_train);
\draw[arr] (test) -- (calib_test);
\draw[arr] (calib_test.east) -- (obs1.west);

\draw[arrdash] (calib_train.south) -- (calib_test.north)
  node[midway, right, annot, xshift=2pt] {item difficulty};

\coordinate (banner) at ([yshift=0.55cm]ann-ct.north);
\node[stlabel] at (banner -| dataset) {\textsc{Input}};
\node[stlabel] at (banner -| filter)  {\textsc{Preprocessing}};
\coordinate (partmid) at ($(train)!0.5!(test)$);
\node[stlabel] at (banner -| partmid) {\textsc{Partition}};
\node[stlabel] at (banner -| calib_train) {\textsc{Training}};
\node[stlabel] at (banner -| predict) {\textsc{Evaluation}};

\begin{scope}[on background layer]
  \node[fit=(banner -| dataset)(ann-data)(ann-pred)(pred1)(ann-ct),
        fill=cBG, rounded corners=8pt, inner sep=10pt] (bgbox) {};
  \coordinate (sep1) at ($(dataset.east)!0.5!(filter.west)$);
  \draw[cGray!20, densely dashed, line width=0.4pt]
    ([yshift=4pt]sep1 |- bgbox.north) -- ([yshift=-4pt]sep1 |- bgbox.south);
  \coordinate (sep2) at ($(filter.east)!0.45!(train.west)$);
  \draw[cGray!20, densely dashed, line width=0.4pt]
    ([yshift=4pt]sep2 |- bgbox.north) -- ([yshift=-4pt]sep2 |- bgbox.south);
  \coordinate (sep3) at ($(train.east)!0.5!(calib_train.west)$);
  \draw[cGray!20, densely dashed, line width=0.4pt]
    ([yshift=4pt]sep3 |- bgbox.north) -- ([yshift=-4pt]sep3 |- bgbox.south);
  \coordinate (sep4) at ($(calib_test.east)!0.5!(predict.west)$);
  \draw[cGray!20, densely dashed, line width=0.4pt]
    ([yshift=4pt]sep4 |- bgbox.north) -- ([yshift=-4pt]sep4 |- bgbox.south);
\end{scope}

\end{tikzpicture}

%% file: sections/3data.tex
\section{A Dataset of Problem-Solving Trajectories}

\begin{figure*}[t]
    \centering
    \includegraphics[width=\textwidth]{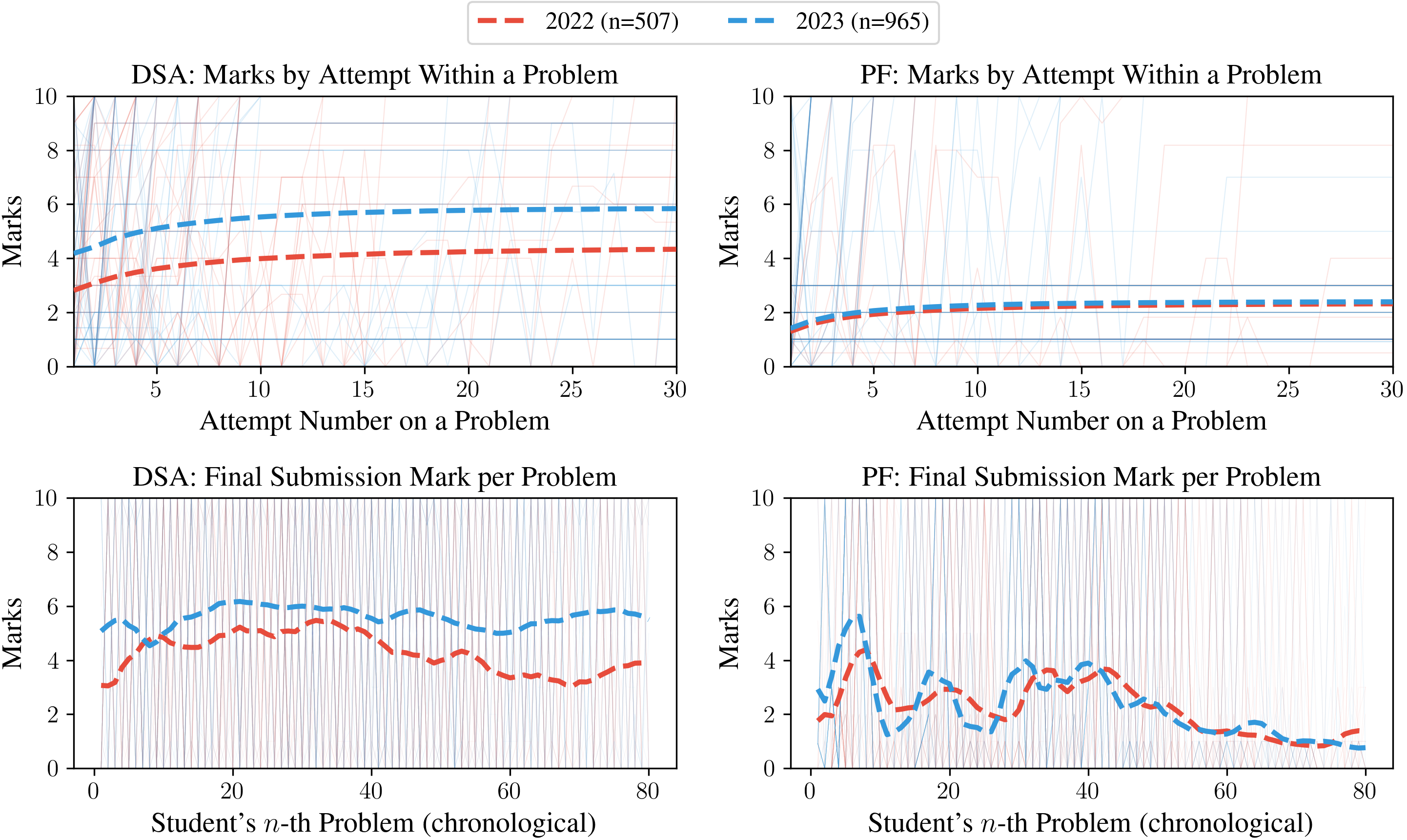}
    \caption{Top: average marks vs. number of attempts for DSA (left) and PF (right), comparing 2022 and 2023 cohorts. Dashed lines show cohort averages across all students and problems; background lines show individual student trajectories. Bottom: final submission scores by chronological problem order. Scores fluctuate across problems with problem difficulty; no monotonic improvement trend emerges.}
    \label{fig:learning-curves}
\end{figure*}

The CodeInsight dataset comprises 3,286 undergraduate students from the Department of Computer Science at VNU-HCM University of Technology (Vietnam) across the 2022 and 2023 academic years. We collected data from two courses: Programming Fundamentals (PF, Spring semester), a first-year course, and Data Structures and Algorithms (DSA, Fall semester), a second-year course that requires PF as a prerequisite. Each course was offered twice, and within each offering, students were divided into cohorts by admission pathway: regular admission (L), higher-tuition track with lower entrance scores (CC), and retaking (DT). The dataset is structured at two distinct levels of granularity: a \textbf{problem} constitutes a discrete programming exercise evaluated via multiple hidden \textbf{test cases}, each yielding a binary outcome. A \textbf{submission} represents a single iterative attempt to solve a given problem, which is subsequently evaluated independently across all associated test cases. Table~\ref{tab:dataset_summary} summarizes the dataset statistics.

\begin{table}[t]
  \centering
  \caption{Summary statistics of the CodeInsight dataset, broken down by course. Students and problems may appear in multiple courses, so per-course counts do not sum to the unique total.}
  \label{tab:dataset_summary}
  \resizebox{\columnwidth}{!}{%
  \begin{tabular}{lrrrr}
    \toprule
    \textbf{Course} & \textbf{Students} & \textbf{Problems} & \textbf{Test Cases} & \textbf{Submissions} \\
    \midrule
    DSA HK231 (Fall '23) & 966 & 227 & 2,194 & 865,912 \\
    DSA HK221 (Fall '22) & 513 & 175 & 1,661 & 456,916 \\
    PF HK232 (Spring '23) & 1,485 & 129 & 1,307 & 854,104 \\
    PF HK222 (Spring '22) & 1,413 & 119 & 1,201 & 897,863 \\
    \bottomrule
    \\[-0.5em]
    \textbf{Unique Total} & 3,286 & 394 & 3,888 & 3,074,795 \\
  \end{tabular}%
  }
\end{table}

To ensure privacy, student identities are systematically anonymized using numeric indices. This anonymization process included the removal of all names and unique personal identifiers, as well as a thorough screening of the submissions to verify the absence of embedded emails or student identification numbers within the code. The platform logs exact submission timestamps alongside sequential student actions, such as starting, prechecking, saving, checking, and finishing attempts, within an interactive coding environment. The finalized dataset encompasses the students' raw source code, exact submission times, admission-track labels, and all evaluation test cases, including those originally designated as private. This dataset remains the copyrighted intellectual property of the Department and is released under a specific departmental license intended strictly for educational purposes upon request. Academic researchers may request access under this departmental license. Approved requesters receive the raw source code of every submission, the full problem statements and code templates, the public and private test cases with per-test-case pass/fail labels for every submission, the grading scripts used to compile and evaluate submissions, and the train/test split indices used in our experiments. Figure~\ref{fig:data-ui} illustrates the platform interface, detailing the problem statement, code editor, and automated feedback mechanisms for both incorrect and fully correct submissions. Additional details regarding the platform architecture and the submission workflow are provided in Appendix~\ref{sec:apd_data}.

The dataset comprises over 3 million programming submissions, capturing rich temporal and behavioral patterns that reflect authentic student engagement. Temporal analysis shows that students primarily work during evening hours (8 PM to 11 PM), with submission activity peaking on Saturdays, demonstrating typical undergraduate study patterns. The median time between consecutive submissions is 59 seconds, indicating rapid iterative development cycles where students frequently test and refine their code in response to automated feedback. Analysis across three behavioral dimensions (time between submissions, edit distance, and total submission count) shows that the majority of students occupy a central cluster characterized by moderate edit distances, short inter-submission intervals, and several hundred total submissions, consistent with this iterative workflow. A smaller set of outliers deviates from this pattern, exhibiting either unusually large edit distances or very long gaps between submissions, which may reflect off-platform work, copy-paste behavior, long interruptions, or other unobserved activity. A detailed breakdown of these behavioral patterns is provided in Appendix~\ref{sec:apd_data}, and example student trajectories with full submission sequences are shown in Appendix~\ref{sec:student_trajectories}.

Figure~\ref{fig:learning-curves} presents learning curves across attempts for both courses, comparing the 2022 and 2023 cohorts. For each student-problem pair, we compute the score (fraction of test cases passed) at each attempt and average across all pairs at each attempt number. Both are foundational courses with largely invariant syllabi across years, so curriculum changes are an unlikely source of the cohort differences. Across both courses, early attempts on a given problem yield the largest score improvements, followed by diminishing returns. However, the year-over-year difference is substantially more pronounced in DSA than in PF, where the 2022 and 2023 cohorts follow nearly identical trajectories.
Since each curve tracks repeated attempts on the same problem, task identity is controlled. Within a fixed time window, to avoid survivorship bias from students who stop attempting early, a student's final score is propagated to all subsequent attempt indices, so the population remains constant across attempts. The resulting curves should be interpreted as descriptive evidence of within-problem improvement rather than causal estimates of learning. In contrast, the bottom row of Figure~\ref{fig:learning-curves} orders each student's problems chronologically by first-attempt timestamp and plots their last-submission score at each position, averaged across students. No consistent upward trend emerges because the average score on a given problem is a function of both student ability and problem difficulty, and these two factors vary independently across the curriculum. Disentangling them requires explicit difficulty estimation, which motivates the parametric models we evaluate in Section~\ref{sec:models}.

At the problem level (Appendix Figure~\ref{fig:problem-patterns}), the number of steps students take to solve a question varies widely, with most problems requiring fewer than 10 attempts but a long tail extending past 20. Last-attempt scores are right-skewed, with a spike near zero where many students do not reach a passing score by their final submission and a long tail spread across higher scores. This distribution highlights the range of problem difficulty in the dataset and the substantial fraction of student-problem interactions that end without a correct solution. In summary, this dataset offers substantial value both quantitatively, through a large student population, high submission volume, and longitudinal tracking, and qualitatively, by capturing fine-grained problem-solving behaviors across different course levels and successive cohorts.

%% file: sections/6Result.tex
\section{Results}

\begin{figure*}[!ht]
  \centering
  \includegraphics[width=\textwidth, keepaspectratio]{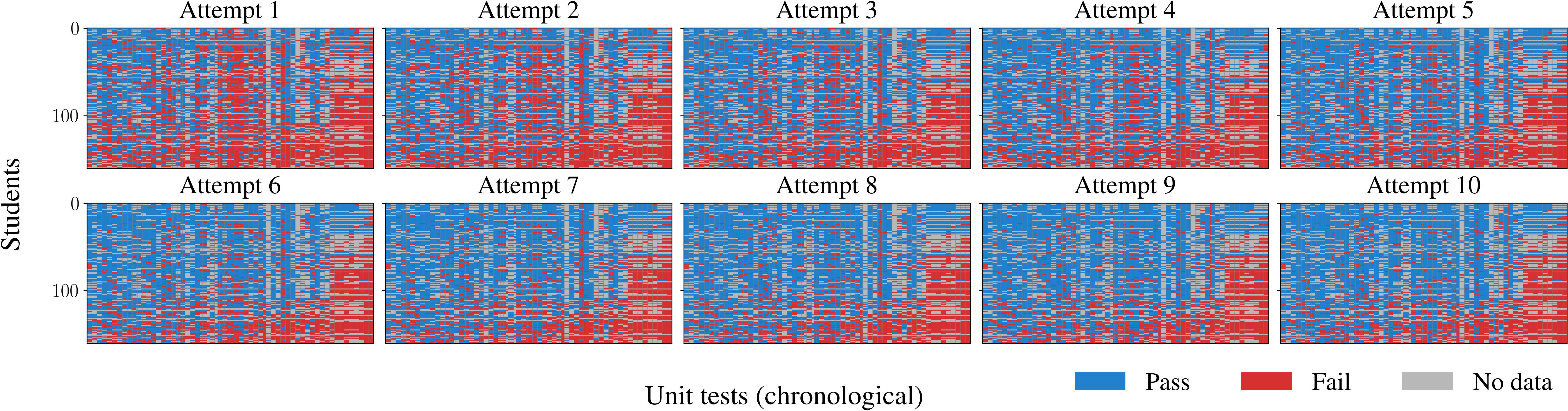}
  \caption{Per-attempt correctness heatmap of students on DSA HK231. Each of the ten panels shows one attempt number. Within each panel, the vertical axis is students (one row per student, sorted by overall pass rate) and the horizontal axis is test cases (one column per test case, ordered chronologically); blue marks a passed test case and red a failed one. The progressive expansion of blue from attempt~1 to attempt~10 shows students mastering additional test cases through repeated practice.}
  \label{fig:attempt_progression}
\end{figure*}

\begin{figure}[!ht]
  \centering
  \includegraphics[width=\columnwidth]{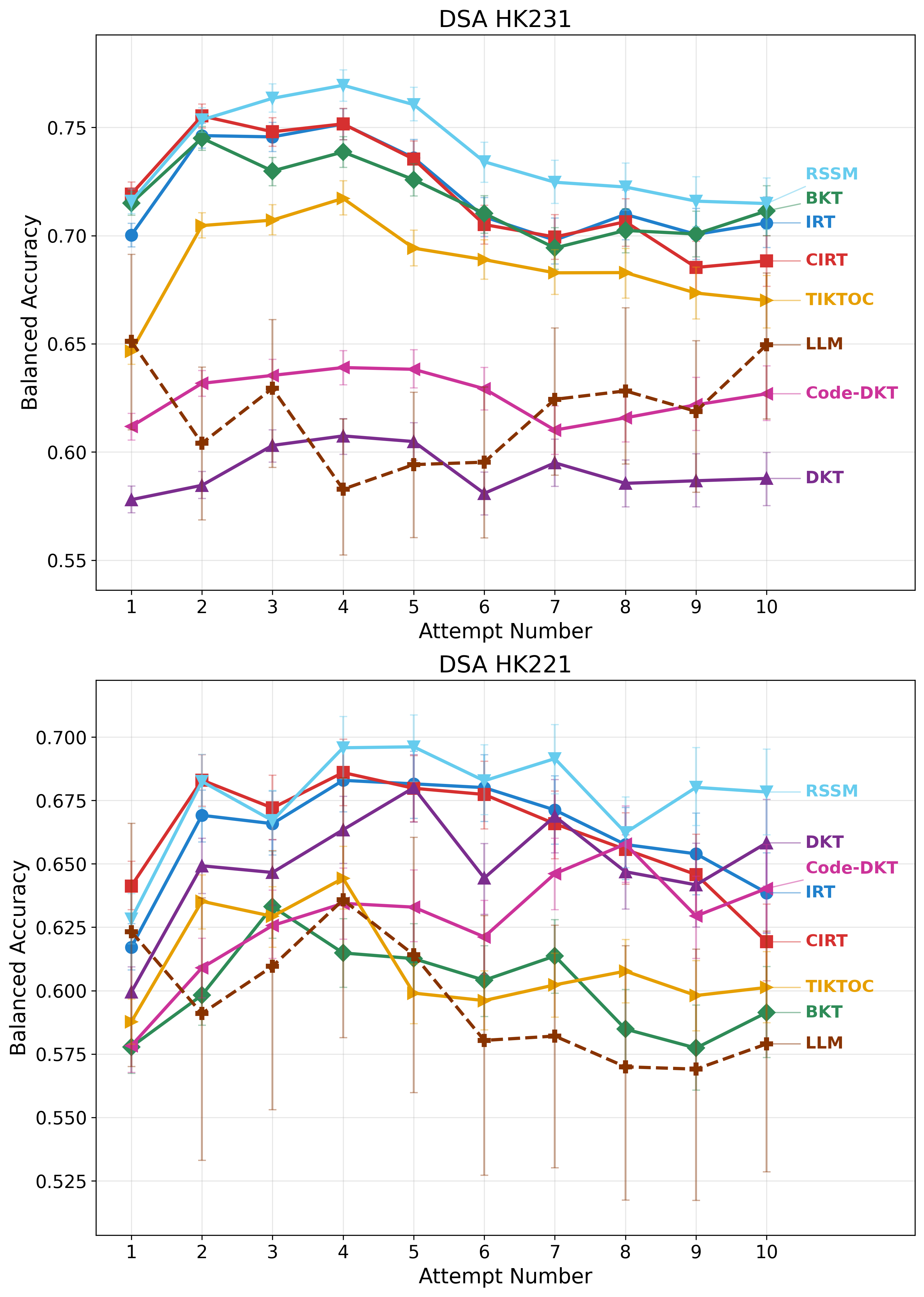}
  \caption{Balanced accuracy at each attempt number across both DSA courses. Top: DSA HK231. Bottom: DSA HK221.}
  \label{fig:accuracy_vs_attempt}
\end{figure}

\begin{figure}[t]
  \centering
  \includegraphics[width=\columnwidth]{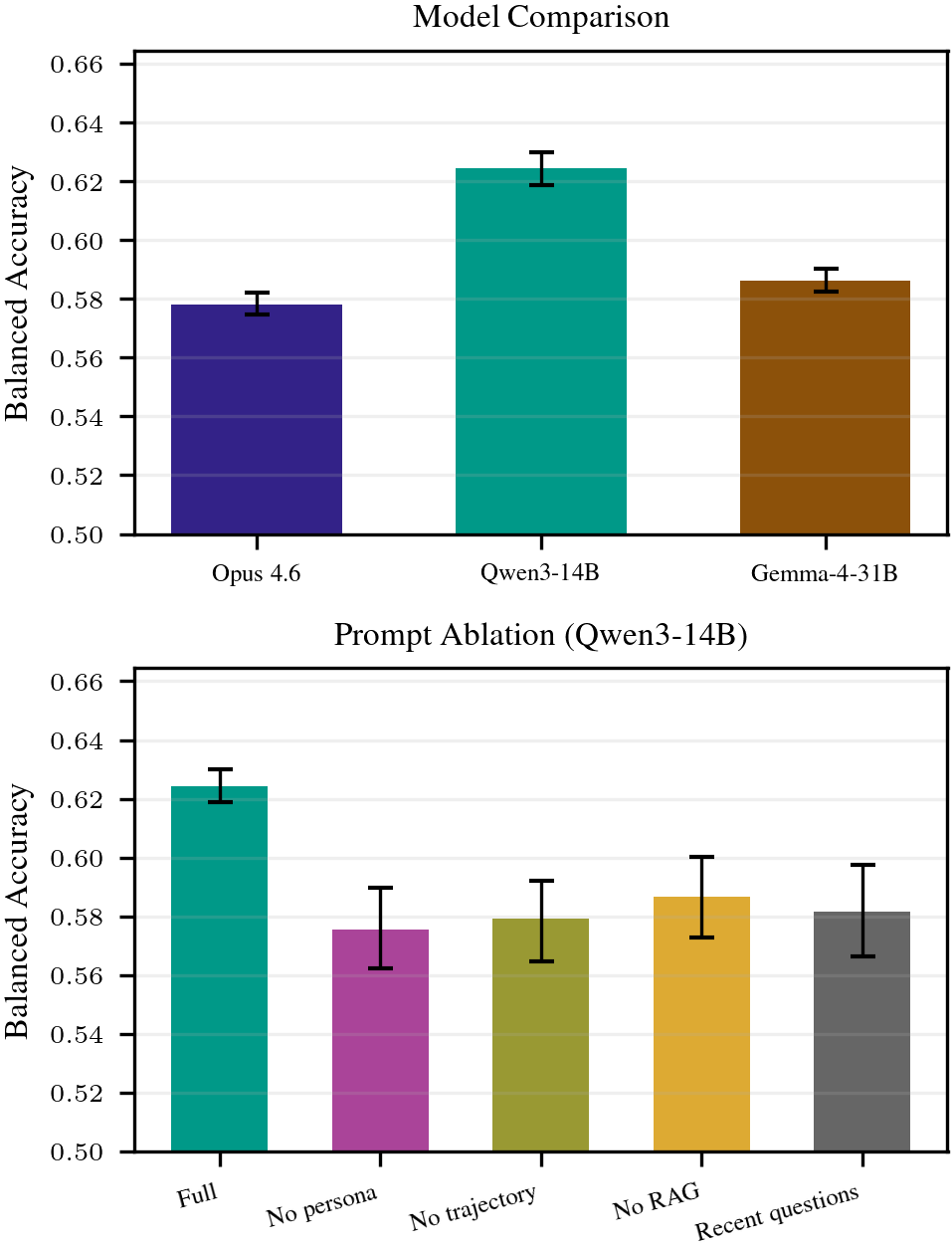}
  \caption{LLM predictor analysis on DSA HK231. Top: model comparison. Qwen3-14B achieves the highest balanced accuracy, outperforming Gemma-4-31B and Opus. Bottom: prompt ablation on Qwen3-14B. The full prompt outperforms all ablated variants.}
  \label{fig:llm_ablation}
\end{figure}

\begin{figure*}[!ht]
  \centering
  \includegraphics[width=\textwidth]{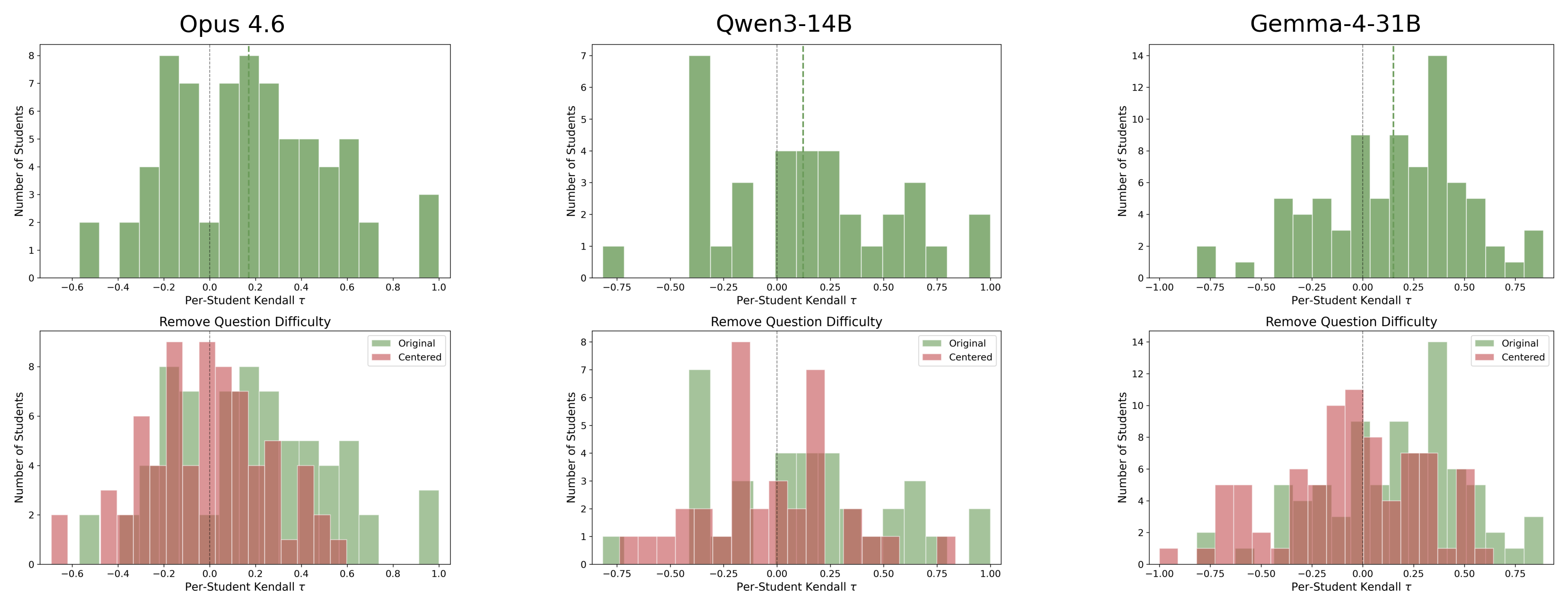}
  \caption{Per-student Kendall $\tau$ between LLM-predicted and real scores across all three LLMs (Opus, Qwen3-14B, Gemma-4-31B). Top row: raw $\tau$ distributions show weak positive means. Bottom row: after subtracting each problem's mean score across students to remove difficulty, the distribution means shift toward zero for all three models, indicating that the predictive signal is driven more by problem-level difficulty than by student-specific modeling.}
  \label{fig:kendall_decomposition}
\end{figure*}

\begin{figure}[!ht]
  \centering
  \includegraphics[width=\columnwidth]{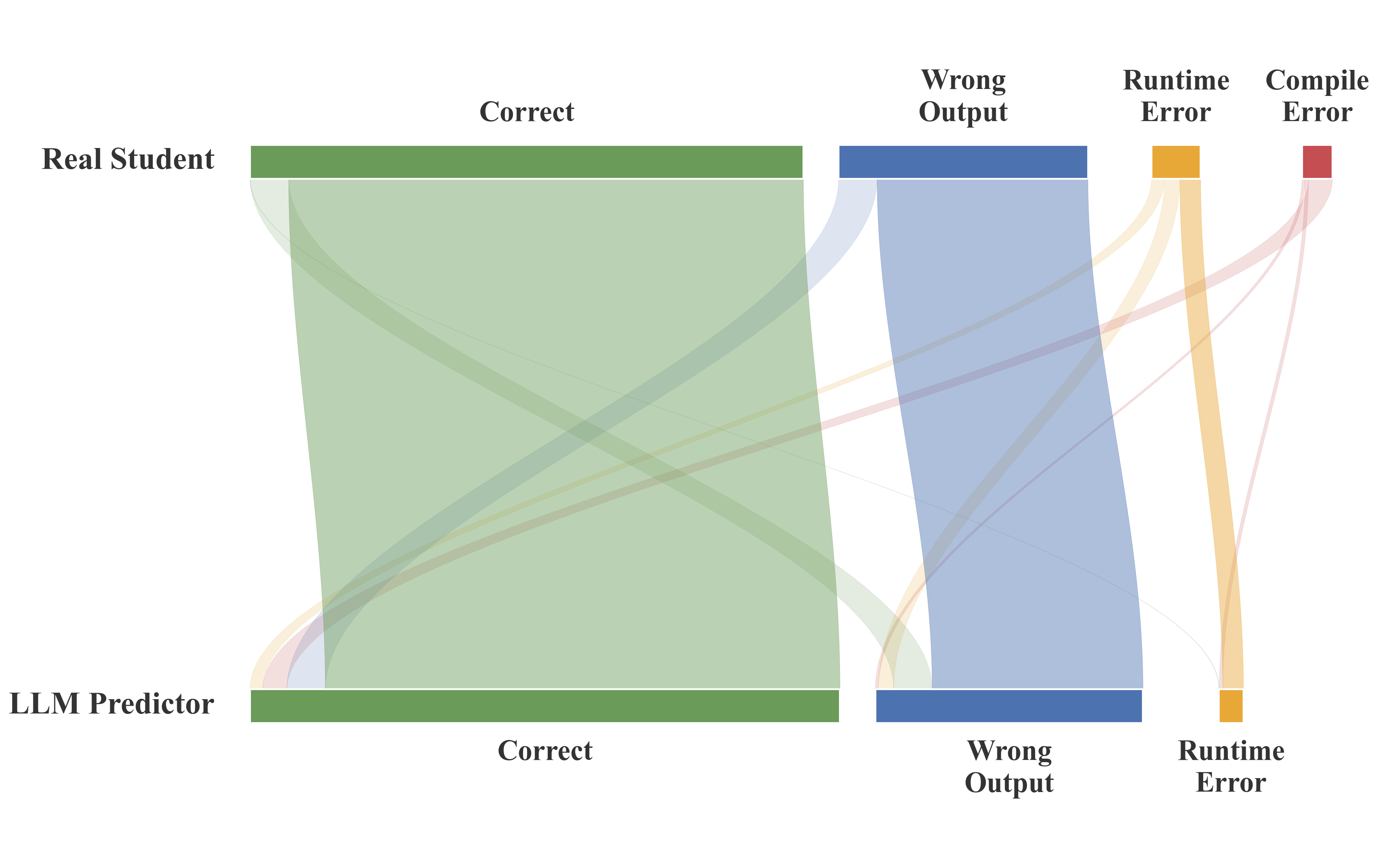}
  \caption{Test-case outcome flow between real students (top) and the LLM predictor (bottom) on matched submissions. Flow width is proportional to the number of test cases sharing that transition. The LLM produces more correct outcomes and fewer runtime and compilation errors than real students, but wrong-output cases transfer at high rates between the two.}
  \label{fig:error_flow}
\end{figure}

\subsection{Experimental Setup}
\label{sec:data_setup}

An overview of our pipeline is presented in Figure~\ref{fig:eval_pipeline}. We filter the dataset to retain trajectories that exhibit iterative problem solving. Problems with pass rates below 10\% or above 90\% are removed, where a problem's pass rate is the fraction of passing test-case outcomes across students' final submissions. Such problems are too difficult or too easy to provide the problem-solving signal this analysis requires: nearly every solver ends at the same outcome, leaving little variation across solvers or attempts for the predictor to explain. Problems attempted by fewer than 25\% of the students are also excluded for insufficient coverage. Appendix~\ref{sec:apd_filtering} reports a sensitivity analysis over these filters: across all variations we test, the model ordering is consistent, and removing the filters entirely inflates absolute scores for every model without changing the comparative conclusions. Figure~\ref{fig:attempt_progression} visualizes the correctness matrix of this data across attempts. Each panel corresponds to one attempt level, with students on the vertical axis and test cases on the horizontal axis. At attempt 1, a substantial portion of the matrix is red, reflecting widespread initial failure. As the attempt number increases, the blue region expands progressively, indicating that students are learning and passing more test cases with repeated practice.

\subsection{Model Comparison}
\label{sec:model_comparison}

We first compare predictive accuracy across all models. Figure~\ref{fig:accuracy_vs_attempt} reports balanced accuracy across attempt numbers on both DSA courses; the RSSM achieves the strongest balanced accuracy and AUC on three of the four courses. The parametric models and the RSSM share a common shape across attempts: balanced accuracy rises over the first few attempts, peaks around attempts two to five, and declines thereafter. The decline is steepest for the parametric models: IRT falls back to its attempt-1 level by attempt 7 on DSA HK231, and CIRT, whose per-problem temporal slope helps at early attempts, drops below IRT at the latest attempts on both courses, indicating that a single linear trend per problem captures limited sequential dynamics. The RSSM declines least, so its advantage over the parametric baselines is concentrated at later attempts. On DSA HK231 it reaches balanced accuracy of 0.71--0.73 at later attempts against 0.69--0.71 for IRT and CIRT. On DSA HK221, where the overall pass rate is lower (29.5\% vs 42.8\% on DSA HK231) and students improve more gradually, it holds 0.66--0.69 against 0.62--0.68. DKT and Code-DKT trail the parametric models on both courses and at every attempt on DSA HK231; on DSA HK221 the gap narrows over the trajectory, and DKT overtakes both parametric models at the last attempt. BKT tracks the parametric models on DSA HK231 but falls below them on DSA HK221. TIKTOC sits between the two groups: on DSA HK231 it runs below the parametric models at every attempt while staying above DKT and Code-DKT, and on DSA HK221 it peaks early (0.64 at attempt 4) and flattens near 0.60 thereafter. The LLM predictor (Qwen3-14B) sits at 0.57--0.65, below the RSSM and the parametric models at nearly every attempt, though it overlaps with DKT and Code-DKT on DSA HK231. The PF courses show a similar shape across attempts, though the ordering shifts: TIKTOC is strongest on PF HK232 and the RSSM on PF HK222. Their outcomes are heavily skewed toward failure (pass rates of 8.4\% and 17.5\%), which makes outcomes easier to predict and inflates absolute scores for every model. Per-attempt metrics for all models and courses are reported in Appendix Table~\ref{tab:per_attempt_metrics}.

Given the LLM's lower accuracy, we investigate what drives this prediction gap. Figure~\ref{fig:llm_ablation} (top) shows the comparison among three LLMs: Qwen3-14B achieves the highest balanced accuracy, outperforming both Gemma-4-31B and Claude Opus 4.6. This ordering is inversely related to coding proficiency: Opus 4.6 is the strongest coder among the three, followed by Gemma-4-31B, with Qwen3-14B the weakest on standard code generation benchmarks such as LiveCodeBench~\cite{jain2024livecodebench}. Stronger coders may over-optimize solutions relative to the errors students actually make, reducing their fidelity as predictors in this context. Figure~\ref{fig:llm_ablation} (bottom) ablates the prompt structure for Qwen3-14B. The full prompt (student persona, prior attempt trajectory, and retrieved examples of the student's approach on similar problems) outperforms all ablated variants, indicating that for this model each component contributes a complementary signal to next-step prediction. A fourth variant replaces similarity-based retrieval with the student's most recently attempted problems and performs comparably to removing retrieval entirely, suggesting that content similarity drives the value of the retrieved examples. 

We next examine whether the LLMs capture solver-specific failure modes or primarily reflect problem-level difficulty. Although the LLM predictors score below the parametric and sequential models on balanced accuracy, they produce actual code at each attempt, enabling a more detailed analysis of how they succeed and fail. Figure~\ref{fig:error_flow} compares test-case outcomes between real students and the LLMs on matched submissions. The LLMs produce correct output on more test cases than the typical student and nearly eliminate runtime errors and compilation errors, consistent with training corpora that consist predominantly of correct, compilable code. For wrong-output cases, the pattern differs: the majority of students' wrong-output cases flow to LLM wrong-output cases, indicating that the LLMs tend to fail on the same test cases where students produce logically incorrect results.

To determine whether this overlap reflects genuine student modeling or simply shared problem-level difficulty, we compute per-student Kendall $\tau$ between LLM-predicted and real scores, then repeat the analysis after subtracting each problem's mean score across all students to isolate the student-specific residual (Figure~\ref{fig:kendall_decomposition}). Across all three LLMs, the same pattern holds: the raw $\tau$ distributions show weak positive means, but after removing the problem-difficulty signal, the distributions' mean moves toward zero. This indicates that for all three models, the predictive signal is driven more by problem-level difficulty than student-specific modeling. A separate test reinforces this finding: when the LLM solves each problem without any student-specific information (Appendix~\ref{sec:prompt_ablation}), its problem-level pass rates correlate strongly ($r = 0.80$) with its pass rates when given student context, while the correlation with real student pass rates is much lower ($r = 0.40$). Together, these analyses establish the central finding of our LLM evaluation: the LLMs do not model individual student ability or strategy beyond problem-level difficulty. The LLM is thus better understood as a generative solver conditioned on student context rather than a faithful student simulator. This finding aligns with recent work by~\citet{ross2025learningmakemistakesmodeling}.

%% file: sections/7Conclusion.tex
\section{Conclusion}

 In this work, we formalize iterative problem-solving as a sequential modeling task, introducing the large-scale CodeInsight dataset to facilitate research into how solvers refine their code over time. By benchmarking diverse models under a shared evaluation framework, we demonstrate that an RSSM capturing discrete latent variables achieves the strongest next-attempt prediction accuracy on three of the four courses, outperforming static parametric baselines with an advantage concentrated at later attempts. Conversely, our investigation into LLMs shows that while they successfully generate analyzable code submissions, they primarily function as generative solvers conditioned on context rather than faithful simulators of human behavioral flaws. 
 
 The practical applications and broader impact of these findings extend to both human learning analytics and the evaluation of autonomous coding agents. For education, modeling persisting errors and shifting strategies could support personalized feedback and adaptive instruction. For the AI and NLP communities, this research highlights a gap in current LLM capabilities, emphasizing the need for future fault-guided training to better capture realistic, student-specific error distributions. Ultimately, by moving beyond binary correctness to model the full arc of iterative refinement, this work provides an empirical foundation for studying systematic failure modes and advancing context-aware problem-solving models.

\section*{Limitations}
Several limitations constrain the scope of our findings. Our dataset is drawn from a single Vietnamese university and covers only C++ programming courses (a syntactically rich and error-prone language relative to Python), and was collected on a single grading platform whose feedback design shapes the trajectories we observe. Whether the observed patterns generalize to other institutions, programming languages, platforms, and course designs remains an open question.

We evaluated three LLMs (Opus, Qwen3-14B, Gemma-4-31B) with frozen parameters and no fine-tuning. Other LLMs may exhibit different strengths or failure modes. Fine-tuning on error-rich student code could improve behavioral fidelity; prior work shows that fault-guided training on buggy code improves code generation quality \citep{fan2025faitfaultawarefinetuningbetter}, suggesting current models are underexposed to realistic error patterns, though we have not tested this in our setting.

Our evaluation is also confined to prediction: accuracy is measured as a modeling capability, and we do not test whether forecasts, if surfaced to instructors or solvers, would improve outcomes. Such an intervention study requires a controlled deployment in a live course and is future work. Score prediction likewise stops short of diagnosis: a model can forecast repeated failure on a test case without identifying the underlying misconception or the help that would resolve it.

Future work should extend the evaluation to additional LLMs and programming languages, further ablate the predictor's design choices, investigate whether other retrieval strategies or constrained decoding can encourage more realistic predictive behavior for the LLMs, and probe the RSSM's discrete latent space for interpretable solver states. These extensions would help establish which findings are specific to this setting and which reflect broader properties of generative models applied to sequential prediction on problem-solving trajectories.

\section*{Ethical Considerations}
The primary ethical considerations in this research center on the curation and distribution of the CodeInsight dataset, which comprises over 3 million code submissions from 3,286 undergraduate students. As this research is performed after the courses are completed, we obtained permission from the Department to curate and use the data. To protect student privacy, all identities were anonymized using numeric indices. This anonymization process included the removal of all names and unique personal identifiers from the data. Additionally, we conducted a thorough screening of the submissions to verify that no embedded emails or student identification numbers remained within the source code. To mitigate the risks of data misuse, the dataset remains the copyrighted intellectual property of the university department and is released under a specific departmental license intended strictly for educational purposes upon request.

\section*{Acknowledgments}
SK is partially supported by NSF 2046795, 2205329, and 2504264, NIH, ARPA-H, the MacArthur Foundation, Good Ventures, Schmidt Sciences, the Hasso Plattner F\"{o}rderstiftung, and Stanford HAI. We acknowledge the Faculty of Computer Science and Engineering at VNU-HCMUT for granting permission to use this data for research. We also thank the Department of Computer Science, the teaching assistants, and students who contributed the questions and submissions that make up this dataset.

%% file: sections/8.10A10.DatasetDetails.tex
\section{Dataset Details}
\label{sec:apd_data}

Throughout the course, students engage in weekly programming assignments that involve a variety of topics, starting with object-oriented programming (OOP), recursion, array lists, and singly linked lists in the first week. The subsequent weeks introduce more advanced data structures and algorithms, including doubly linked lists, stacks, queues, sorting algorithms, binary trees, AVL trees, and search algorithms, culminating in topics like hash functions and graph theory by the final week. Each week begins with an exam covering the previous week's material.

Prechecks allow students to test their code against a public set of test cases, providing feedback on whether their code compiles and passes basic tests. Students see the input, output, and feedback on their code's performance. In contrast, submissions are evaluated against a private set of test cases that assess the robustness of their solutions, with more challenging test cases and potential penalties for multiple submissions. Here, students only see a numerical evaluation score.

The course has six weeks of lab sections; each week covers several topics, and each topic includes several questions. Students will do exercises by navigating to each topic and attempting the questions inside. They may try multiple times. For each question, they receive a problem statement and a code editor. After completing their code, they can press ``Submit'' to check if they have passed the test cases. If they fail the test cases, they can edit their code and resubmit multiple times. The system will save each submission. Students will receive scores based on the number of test cases passed. The score for a topic is the sum of the scores for each related question. Whenever students submit their answers, they receive information about which test cases have passed and which have not. Using this feedback, students should be able to learn to fix their code to perform better, thereby improving with each submission.

\begin{figure}[t]
    \centering
    \includegraphics[width=\columnwidth]{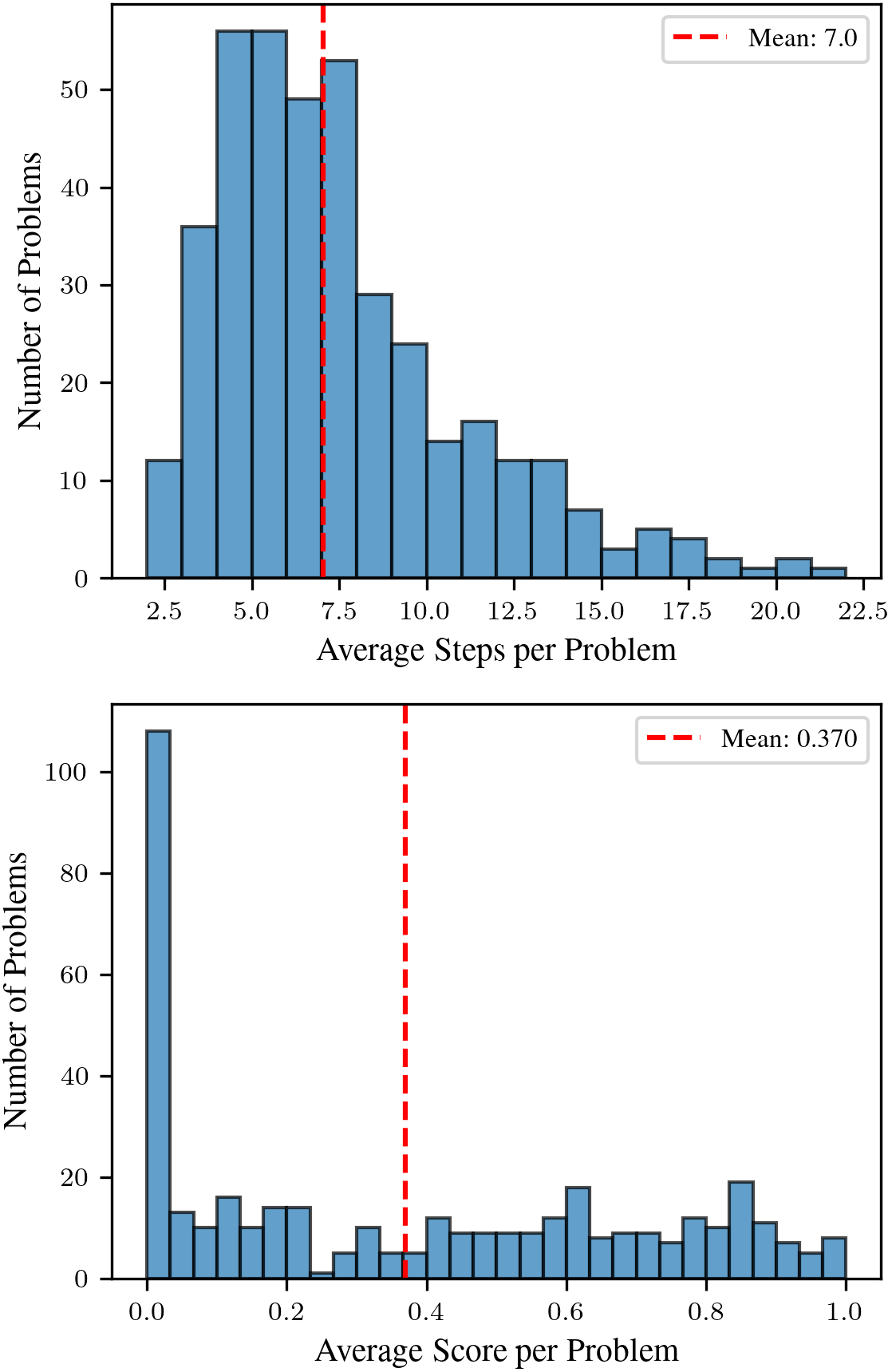}
    \caption{Problem-level distributions across all courses. (Top) Average number of attempts per problem. (Bottom) Average last-attempt score per problem.}
    \label{fig:problem-patterns}
\end{figure}

\subsection{Student Behavior Patterns}

To understand the relationship between student engagement patterns and learning outcomes, we analyzed submission pacing behavior across all students in the dataset. The analysis shows that students who make more submissions per problem tend to achieve higher final scores, demonstrating that persistent engagement with iterative refinement is associated with better outcomes. The median student makes 455 submissions across all problems (mean: 610 submissions), reflecting sustained engagement with the course material throughout the semester. However, the relationship between pause duration (median time between consecutive submissions) and performance is more nuanced, with successful students exhibiting a range of pacing strategies.

Examination of pause time distributions shows that students operate within relatively compressed time scales. The median pause time between consecutive submissions is 59 seconds, with a mean of 2.7 minutes. This pattern suggests that students rely heavily on the automated feedback system to guide incremental code improvements. The predominance of short pause times indicates a trial-and-error approach where students make frequent small adjustments in response to test results.

\begin{figure*}[t]
    \centering
    \includegraphics[width=\textwidth]{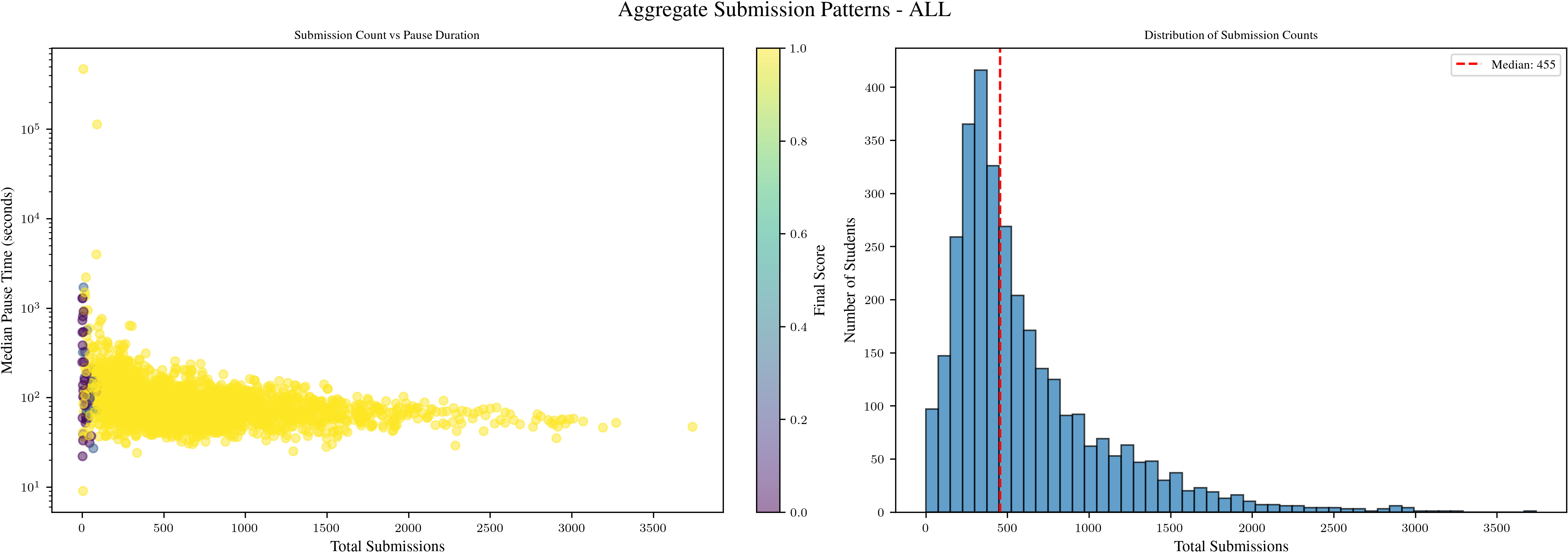}
    \caption{Aggregate submission pacing patterns across all students and courses. (Left) Scatter plot of total submissions vs. median pause time, colored by final score. (Right) Distribution of total submissions per student, with the median (455) marked by the dashed red line.}
    \label{fig:submission-patterns}
\end{figure*}

Analysis of high-volume submitters shows varied performance outcomes (Figure~\ref{fig:submission-patterns}): while some achieve perfect or near-perfect scores through persistent iteration, others continue to struggle despite numerous attempts. This heterogeneity indicates that submission volume alone is not a reliable predictor of success; the quality of revisions and the student's ability to interpret and act on feedback are equally important factors.

\begin{figure*}[t]
    \centering
    \includegraphics[width=\textwidth]{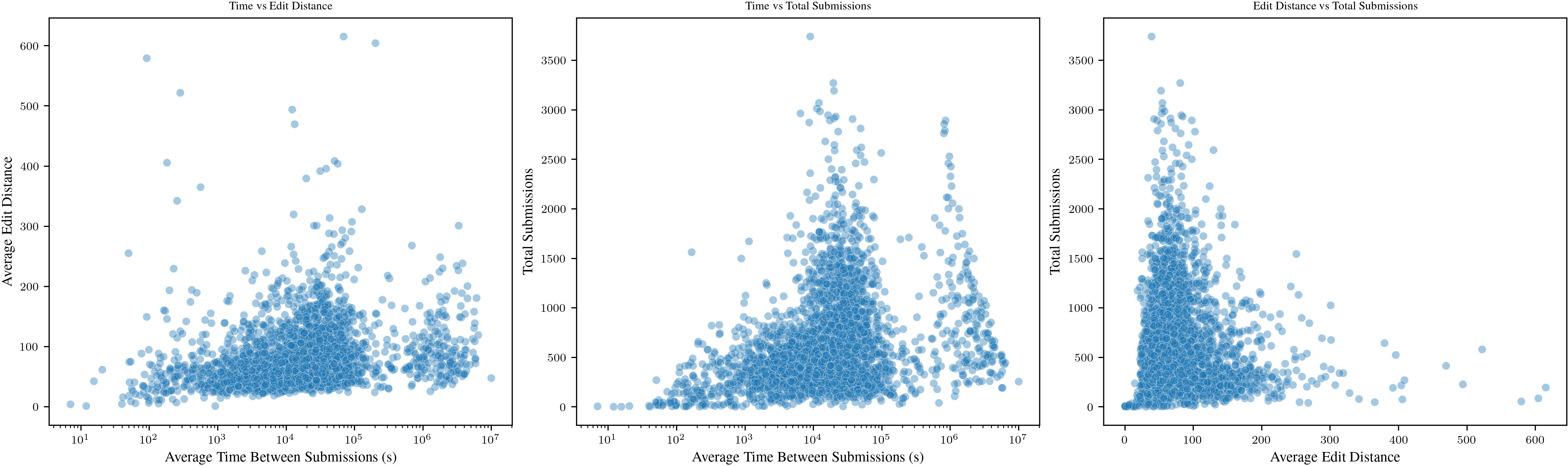}
    \caption{Student coding behavior across all courses. From left to right: average time between submissions vs. average edit distance; average time between submissions vs. total submissions; and average edit distance vs. total submissions.}
    \label{fig:student-behavior}
\end{figure*}

Figure~\ref{fig:student-behavior} shows pairwise relationships among three behavioral features: average time between submissions, average edit distance (Levenshtein distance between consecutive code submissions), and total number of submissions. The plots show that most students cluster in a region of moderate edit distances and submission counts, with a long tail of outliers exhibiting either very high edit distances or very long inter-submission intervals.

\subsection{Course Characteristics and Trajectory Quality}
\label{sec:course_comparison}

\begin{figure*}[!htbp]
    \centering
    \includegraphics[width=\textwidth]{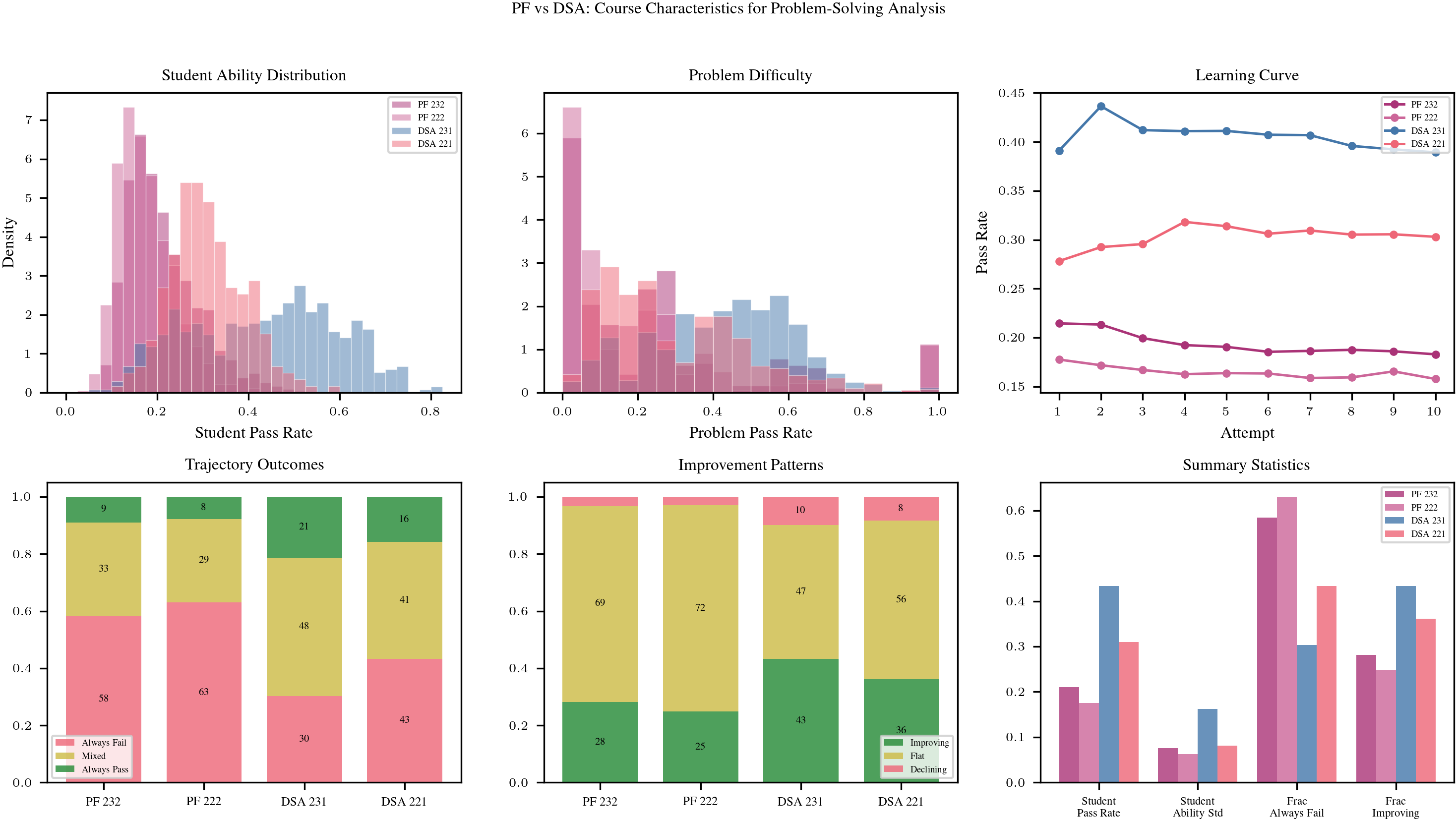}
    \caption{Course characteristics across all four offerings. Top row: solver ability distribution, problem difficulty distribution, and pass rate by attempt number. Bottom row: trajectory outcome breakdown (always-fail / mixed / always-pass), improvement pattern breakdown (improving / flat / declining), and summary statistics. PF courses are dominated by always-fail trajectories and flat dynamics, while DSA courses show more diverse outcomes and gradual improvement.}
    \label{fig:pf_vs_dsa}
\end{figure*}

Figure~\ref{fig:pf_vs_dsa} compares the four course offerings across six dimensions. The DSA courses exhibit substantially richer problem-solving trajectories than the PF courses. Table~\ref{tab:course_comparison} summarizes the key differences.

\begin{table}[!htbp]
  \centering
  \small
  \caption{Trajectory characteristics across courses. PF courses are dominated by floor effects and flat trajectories, while DSA courses show meaningful improvement across attempts.}
  \label{tab:course_comparison}
  \setlength{\tabcolsep}{3pt}
  \resizebox{\columnwidth}{!}{%
  \begin{tabular}{lcccc}
    \toprule
    & PF 232 & PF 222 & DSA 231 & DSA 221 \\
    \midrule
    Mean pass rate & 0.08 & 0.18 & 0.43 & 0.31 \\
    Always-fail trajectories & 58\% & 63\% & 30\% & 43\% \\
    Improving trajectories & 28\% & 25\% & 43\% & 36\% \\
    Flat trajectories & 69\% & 72\% & 47\% & 56\% \\
    Pass rate $\Delta$ (att 1$\to$10) & $-$3.2pp & $-$2.0pp & $-$0.2pp & $+$2.5pp \\
    \bottomrule
  \end{tabular}}
\end{table}

In the PF courses, 58--63\% of multi-attempt trajectories are always-fail: the solver never passes a single test case across all attempts. Overlapping with this group, 69--72\% of trajectories show no meaningful change between their first and second halves. As described in the main text, Figure~\ref{fig:learning-curves} accounts for survivorship bias by applying last-observation-carried-forward, which propagates each solver's most recent score to subsequent attempt indices. When this correction is removed and only solvers who actually made each attempt are counted, PF pass rates decline across attempts ($-$2 to $-$3 percentage points from attempt 1 to 10), indicating that solvers who persist to later attempts are predominantly those who are stuck rather than those who are improving.

The DSA courses, by contrast, show genuine improvement across attempts, though the two offerings differ in how much sequential signal they provide. DSA HK221 has a lower first-attempt pass rate (27.8\% vs 39.1\%), meaning solvers start further from the solution and have more room to improve. Its pass rate trend is the only positive one across all four courses ($+$2.5pp from attempt 1 to 10). Although DSA HK221 has fewer improving trajectories overall (36\% vs 43\% on DSA HK231), the trajectories that do improve show a clearer, more gradual signal. DSA HK221 also has longer attempt sequences (mean 3.79 vs 3.18 per problem) and a higher fraction of problems with 5+ attempts (32\% vs 25\%), giving a recurrent model more sequential context to work with. DSA HK231 has a higher base pass rate and a flatter trend across attempts. These differences are reflected in the main results (Figure~\ref{fig:accuracy_vs_attempt}): the RSSM leads the parametric models on both DSA courses, with its advantage concentrated at later attempts.

This gradient in trajectory quality (DSA HK221 with the strongest sequential signal, then DSA HK231, then PF with no sequential signal) explains why the main results focus on the DSA courses.

%% file: sections/8.7A7.StudentTrajectories.tex
\section{Examples of Student Learning Trajectories}
\label{sec:student_trajectories}
The following examples illustrate typical student learning trajectories from the CodeInsight dataset. These trajectories show how students iteratively debug and refine their code in response to automated test feedback, exhibiting the behavioral patterns (rapid iteration, incremental edits, and varying debugging strategies) that the knowledge-tracing and LLM-based models aim to predict.

Attempt 1 - 25/09/23, 07:21:06: Started
Attempt result: Not complete. Marks: 0.

Attempt 2 - 25/09/23, 08:35:00: Submit:

\begin{lstlisting}[language=C, breaklines=true, basicstyle=\footnotesize\ttfamily]
    int buyCar(int *nums, int length, int k) {
        int num = 0;
        while (k > 0) {
            int index = 0;
            int minPrice = nums[0];
            for (int i = 0; i < length; i++) {
                if (minPrice > nums[i]) {
                    minPrice = nums[i];
                    index = i;
                }
            }
            k -= minPrice;
            num++; nums[index] = 999999;
        }
        return num - 1;
    }
\end{lstlisting}

Attempt result: Partially correct. Marks: 0.40.

Attempt 3 - 25/09/23, 08:35:16: Submit:
\begin{lstlisting}[language=C, breaklines=true, basicstyle=\footnotesize\ttfamily]
    int buyCar(int *nums, int length, int k) {
        int num = 0;
        while (k $>$ 0) {
            int index = 0;
            int minPrice = nums[0];
            for (int i = 0; i $<$ length; i++) {
                if (minPrice $>$ nums[i]) minPrice = nums[I];
                index = I;
            }
            k -= minPrice;
            num++; nums[index] = 999999999;
        }
        return num - 1;
    }
\end{lstlisting}

Attempt result: Partially correct. Marks: 0.40.

Attempt 4 - 25/09/23, 08:35:36: Submit:
\begin{lstlisting}[language=C, breaklines=true, basicstyle=\footnotesize\ttfamily]
    int buyCar(int *nums, int length, int k) {
        int num = 0;
        while (k $>$ 0) {
            int index = 0;
            int minPrice = nums[0];
            for (int i = 0; i $<$ length; i++) {
                if (minPrice $>$ nums[i]) minPrice = nums[I];
                index = I;
            }
            k -= minPrice; num++;
            nums[index] = 2147483647;
        }
        return num - 1;
    }
\end{lstlisting}

Attempt result: Partially correct. Marks: 0.40.

Attempt 5 - 25/09/23, 08:44:40: Submit:
\begin{lstlisting}[language=C, breaklines=true, basicstyle=\footnotesize\ttfamily]
    int buyCar(int *nums, int length, int k) {
        int num = 0;
        while (k $>=$ 0) {
            int index = 0; int minPrice = nums[0];
            for (int i = 0; i $<$ length; i++) {
                if (minPrice $>$ nums[i]) {
                    minPrice = nums[I];
                    index = I;
                }
            }
            k -= minPrice; num++; nums[index] = 999999;
        }
        return num - 1;
    }
\end{lstlisting}

Attempt result: Correct. Marks: 1.00.

Attempt 6 - 27/09/23, 17:33:21: Attempt finished submitting.

The following trajectory shows a student who submits a complete solution 9 seconds after starting the attempt, suggesting the code was written or prepared before the recorded attempt began:

Attempt 1 - 29/09/23, 13:26:13: Started.
Attempt result: Not complete. Marks: 0.

Attempt 2 - 29/09/23, 13:26:22: Submit:
\begin{lstlisting}[language=C, breaklines=true, basicstyle=\footnotesize\ttfamily]
    int mins[1000]{}, maxs[1000]{}, mine[1000]{}, mmaxe[1000]{};
    int minimumAmplitude(vector<int>& nums, int k) {
        int ans = 1e9;
        int n = nums.size();
        int* mins = new int[n] {};
        int* mine = new int[n] {};
        int* maxs = new int[n] {};
        int* maxe = new int[n] {};
        mins[0] = nums[0]; maxs[0] = nums[0]; mine[n - 1] = nums[n - 1];
        maxe[n - 1] = nums[n - 1];
        for (int i = 1; i < n; i++) {
            mins[i] = min(mins[i - 1], nums[i]);
            maxs[i] = max(maxs[i - 1], nums[i]);
        }
        for (int i = n-2; i >= 0; i--) {
            mine[i] = min(mine[i + 1], nums[i]);
            maxe[i] = max(maxe[i + 1], nums[i]);
        }
        int min_arr = 0;
        int max_arr = 0;
        for (int i = 0; i < n-k; i++) {
            if(i!= 0) min_arr = min(mins[i-1], mine[i+k]);
            else min_arr = mine[i + k];
            if (i + k < n - 1) max_arr = max(maxs[i - 1], maxe[i + k]);
            else max_arr = maxs[i]; ans = min(ans, max_arr - min_arr);
        }
        return ans;
    }
\end{lstlisting}

Attempt result: Correct. Marks: 1.00.

Attempt 4 - 29/09/23, 13:29:58: Attempt finished submitting.

An example for using precheck to check the response before officially submitting:

Attempt 1 - 9/10/23, 09:33:20: Started
Attempt result: Not complete. Marks: 0.

Attempt 2 - 21/10/23, 10:06:07: Prechecked:

\begin{lstlisting}[language=C, breaklines=true, basicstyle=\footnotesize\ttfamily]
    void push(T item) {
        // TODO: Push new element into the end of the queue
        list->add(item);
    }
    T pop() {
        // TODO: Remove an element in the head of the queue
        T ans =list->get(0);
        list->removeAt(0);
        return ans;
    }
    T top() {
        // TODO: Get value of the element in the head of the queue
        return list->get(0);
    }
    bool empty() {
        // TODO: Determine if the queue is empty
        return list->size() == 0;
    }
    int size() {
        // TODO: Get the size of the queue
        return list->size();
    }
    void clear() {
        // TODO: Clear all elements of the queue
        list->clear();
    }
\end{lstlisting}

Attempt result: Precheck results. Marks: 0.

Attempt 3 - 21/10/23, 10:06:58: Prechecked:

\begin{lstlisting}[language=C, breaklines=true, basicstyle=\footnotesize\ttfamily]
    void push(T item) {
        // TODO: Push new element into the end of the queue
        list->add(item);
    }
    T pop() {
        // TODO: Remove an element in the head of the queue
        T ans =list->get(0);
        list->removeAt(0);
        return ans;
    }
    T top() {
        // TODO: Get value of the element in the head of the queue
        return list->get(0);
    }
    bool empty() {
        // TODO: Determine if the queue is empty
        return list.size() == 0;
    }
    int size() {
        // TODO: Get the size of the queue
        return list.size();
    }
    void clear() {
        // TODO: Clear all elements of the queue
        list->clear();
    }
\end{lstlisting}

Attempt result: Precheck results. Marks: 0.

Attempt 4 - 21/10/23, 10:07:12: Submit:

\begin{lstlisting}[language=C, breaklines=true, basicstyle=\footnotesize\ttfamily]
    void push(T item) {
        // TODO: Push new element into the end of the queue
        list->add(item);
    }
    T pop() {
        // TODO: Remove an element in the head of the queue
        T ans =list->get(0);
        list->removeAt(0);
        return ans;
    }
    T top() {
        // TODO: Get value of the element in the head of the queue
        return list->get(0);
    }
    bool empty() {
        // TODO: Determine if the queue is empty
        return list.size() == 0;
    }
    int size() {
        // TODO: Get the size of the queue
        return list.size();
    }
    void clear() {
        // TODO: Clear all elements of the queue
        list->clear();
    }
\end{lstlisting}

Attempt result: Incorrect. Marks: 0.00.

Attempt 5 - 21/10/23, 10:07:32: Prechecked:

\begin{lstlisting}[language=C, breaklines=true, basicstyle=\footnotesize\ttfamily]
    void push(T item) {
        // TODO: Push new element into the end of the queue
        list.add(item);
    } T pop() {
        // TODO: Remove an element in the head of the queue
        T ans =list->get(0);
        list.removeAt(0);
        return ans;
    }
    T top() {
        // TODO: Get value of the element in the head of the queue
        return list->get(0);
    }
    bool empty() {
        // TODO: Determine if the queue is empty
        return list.size() == 0;
    }
    int size() {
        // TODO: Get the size of the queue
        return list.size();
    }
    void clear() {
        // TODO: Clear all elements of the queue
        list.clear();
    }
\end{lstlisting}

Attempt result: Precheck results. Marks: 0.00.

Attempt 6 - 21/10/23, 10:07:36: Submit:

\begin{lstlisting}[language=C, breaklines=true, basicstyle=\footnotesize\ttfamily]
    void push(T item) {
        // TODO: Push new element into the end of the queue
        list.add(item);
    }
    T pop() {
        // TODO: Remove an element in the head of the queue
        T ans =list->get(0);
        list.removeAt(0);
        return ans;
    }
    T top() {
        // TODO: Get value of the element in the head of the queue
        return list->get(0);
    }
    bool empty() {
        // TODO: Determine if the queue is empty
        return list.size() == 0;
    }
    int size() {
        // TODO: Get the size of the queue
        return list.size();
    }
    void clear() {
        // TODO: Clear all elements of the queue
        list.clear();
    }
\end{lstlisting}

Attempt result: Incorrect. Marks: 0.00.

Attempt 7 - 21/10/23, 10:07:49: Submit:

\begin{lstlisting}[language=C, breaklines=true, basicstyle=\footnotesize\ttfamily]
    void push(T item) {
        // TODO: Push new element into the end of the queue
        list.add(item);
    }
    T pop() {
        // TODO: Remove an element in the head of the queue
        T ans =list.get(0);
        list.removeAt(0);
        return ans;
    }
    T top() {
        // TODO: Get value of the element in the head of the queue
        return list.get(0);
    }
    bool empty() {
        // TODO: Determine if the queue is empty
        return list.size() == 0;
    }
    int size() {
        // TODO: Get the size of the queue
        return list.size();
    }
    void clear() {
        // TODO: Clear all elements of the queue
        list.clear();
    }
\end{lstlisting}

Attempt result: Correct. Marks: 1.00.

Attempt 8 - 22/10/23, 20:40:43: Attempt finished submitting.

%% file: sections/8.11A11.TrainingDetails.tex
\section{Model Details}
\label{sec:training_details}

All models operate on the same data and evaluation protocol. Students are split into train (55\%), validation (15\%), and test (30\%) sets using a fixed random seed of 42; the validation students are used only for checkpoint selection and early stopping where a model requires them, and the test set is evaluated once per model. The atomic prediction unit is the individual test case: each problem contains multiple test cases, and each test case is modeled as an independent binary (pass/fail) outcome. Missing attempts are represented with a sentinel value of $-1$ in the three-dimensional correctness tensor (students $\times$ test cases $\times$ attempts) and are masked out during both training and evaluation. Attempts are capped at 10 per student-problem pair. Table~\ref{tab:hyperparams} summarizes the key hyperparameters.

\begin{table}[!htbp]
\centering
\setlength{\tabcolsep}{3pt}
\footnotesize
\resizebox{\columnwidth}{!}{%
\begin{tabular}{@{}lccccccc@{}}
\toprule
 & IRT & CIRT & BKT & DKT & Code-DKT & TIKTOC & RSSM \\
\midrule
Optimizer & Adam & Adam & EM & Adam & Adam & AdamW & Adam \\
Learn.\ rate & 0.01 & 0.01 & -- & 0.001 & 0.001 & 1e-5 / 5e-4 & 3e-4 \\
Epochs & 3000 & 3000 & 50 it. & 200 & 200 & 5 & 200 \\
Batch size & full & full & -- & 100 & 100 & 32 & 64 \\
Regular. & -- & -- & clip & dr.\ 0.5 & dr.\ 0.2 & dr.\ 0.05 & dr.\ 0.1 \\
Early stop & no & no & no & no & no & yes & yes \\
Hidden dim & -- & -- & -- & 64 & 64 & 256 & 512 \\
\bottomrule
\end{tabular}}
\caption{Hyperparameter summary. ``full'': full-batch; ``clip'': parameters clipped to $[0.01, 0.99]$; ``dr.'': dropout; ``it.'': EM iterations.}
\label{tab:hyperparams}
\end{table}

\subsection{IRT}
\label{sec:apd_irt}

A one-parameter logistic (1PL) Rasch model that predicts correctness as $P(\text{correct}) = \sigma(\theta_s - b_j)$, where $\theta_s$ is a per-solver parameter and $b_j$ is a per-test-case difficulty parameter. Both are estimated via maximum likelihood. IRT is static: it assigns each solver a single parameter with no temporal component.

\begin{figure}[!htbp]
  \centering
  \includegraphics[width=\columnwidth]{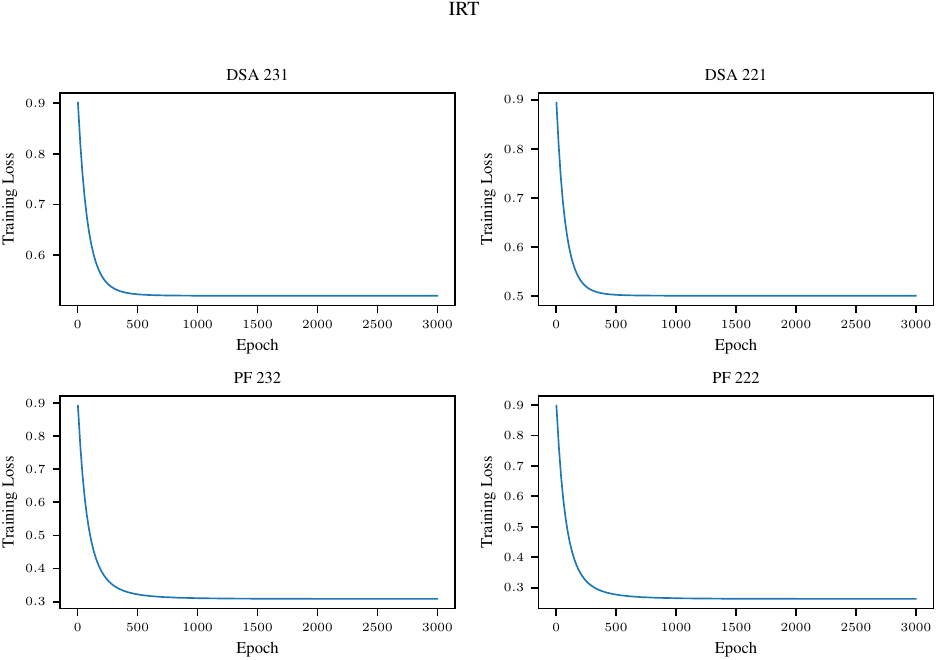}
  \caption{IRT training loss (negative log-likelihood) across epochs for all four course offerings. Loss converges within the first few hundred of 3,000 epochs.}
  \label{fig:irt_loss}
\end{figure}

Training uses Adam with binary cross-entropy loss over all valid observations in a single batch for 3,000 epochs with a learning rate of 0.01. No regularization or early stopping is applied. In the student-split setting, test-case parameters ($b_j$) are first estimated from train students, then frozen. Test student parameters ($\theta_s$) are calibrated on the calibration window (weeks 1--3; predictions cover weeks 4 and later) over 1,000 additional epochs. Training loss curves are shown in Figure~\ref{fig:irt_loss}.

\subsection{CIRT}
\label{sec:apd_cirt}

Continuous IRT extends IRT with a per-problem temporal slope:
\begin{equation}
  P(\text{correct}) = \sigma\!\left(\theta_s - z_j - \lambda_j \cdot t\right)
\end{equation}
where $z_j$ is the baseline difficulty of test case $j$, $\lambda_j$ is a problem-specific difficulty slope, and $t$ is the solver's attempt index. The slope $\lambda_j$ is unconstrained: a negative value indicates that effective difficulty decreases with practice, while a positive value captures diminishing returns from repeated attempts. All parameters are fitted jointly via maximum likelihood.

\begin{figure}[!htbp]
  \centering
  \includegraphics[width=\columnwidth]{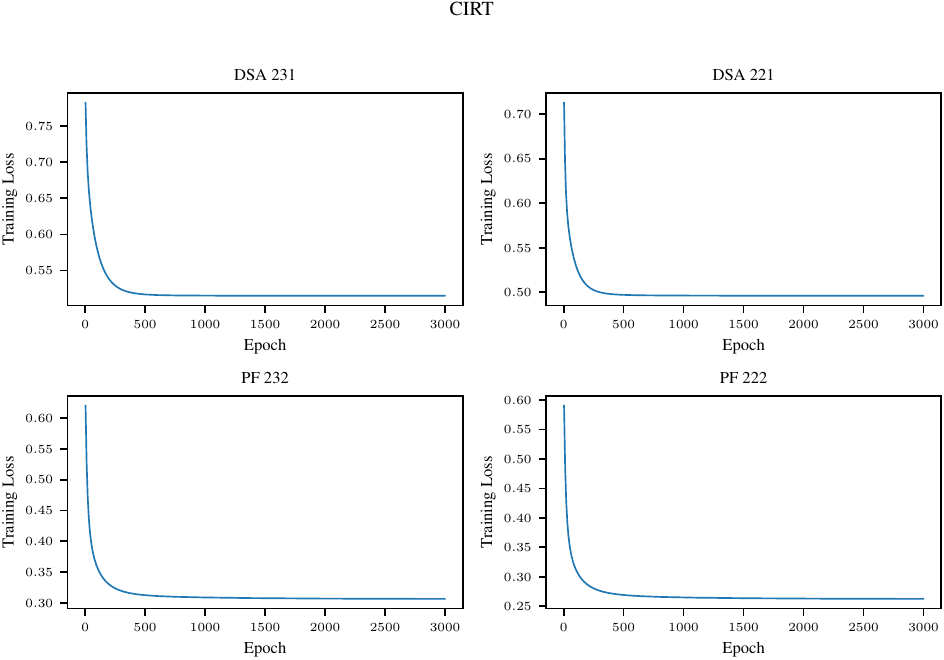}
  \caption{CIRT training loss across epochs. Convergence behavior is similar to IRT, with the additional temporal slope parameter producing comparable final loss values.}
  \label{fig:cirt_loss}
\end{figure}

Training follows the same protocol as IRT: Adam with learning rate 0.01 for 3,000 epochs, full-batch. CIRT adds a per-test-case temporal slope $\lambda_j$ with the attempt index $t$ normalized over $[1, T]$. Training loss curves are shown in Figure~\ref{fig:cirt_loss}.

\subsection{BKT}
\label{sec:apd_bkt}

Bayesian Knowledge Tracing, a hidden Markov model, models each test case as a hidden Markov process with a binary latent state and four parameters: the prior probability of mastery $P(L_0)$, the per-attempt transition probability $P(T)$, the guess probability $P(G)$, and the slip probability $P(S)$.

Each test case is fitted independently via expectation-maximization over all student sequences. The algorithm runs for 50 iterations with fixed initialization ($P(L_0){=}0.3$, $P(T){=}0.1$, $P(G){=}0.2$, $P(S){=}0.1$). After each M-step, all four parameters are clipped to $[0.01, 0.99]$. The forward-backward pass operates on variable-length sequences per student, with missing attempts excluded before fitting. BKT does not produce a training loss curve comparable to the gradient-based models.

\subsection{DKT}
\label{sec:apd_dkt}

Deep Knowledge Tracing, an RNN, processes the sequence of a solver's (test case, outcome) interactions, truncated to the most recent 600, through a single-layer LSTM with hidden dimension 64, followed by dropout (0.5) on the hidden state and a linear readout producing a vector of predicted correctness probabilities at each timestep.

\begin{figure}[!htbp]
  \centering
  \includegraphics[width=\columnwidth]{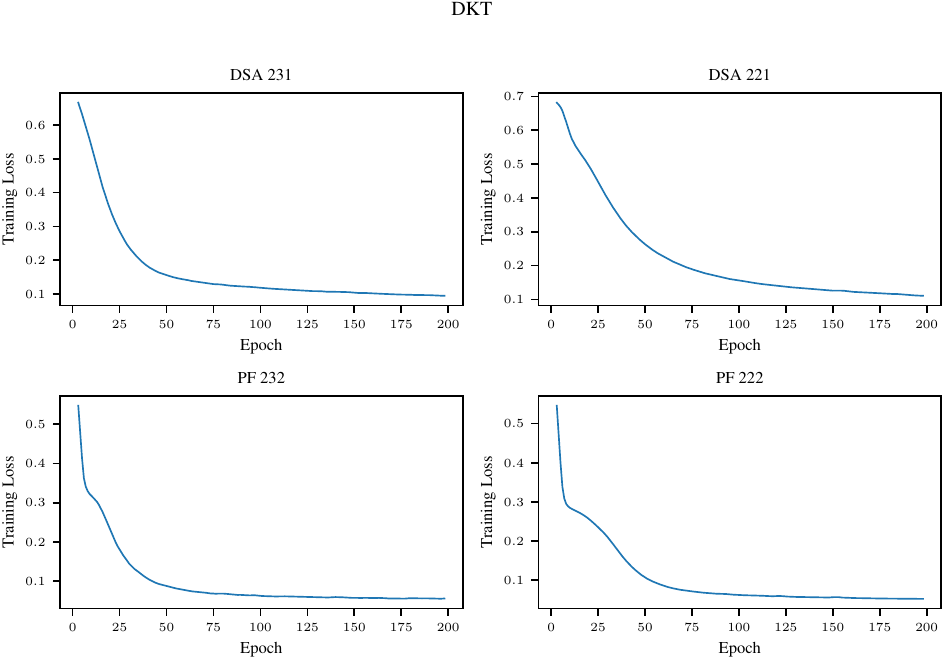}
  \caption{DKT training loss across 200 epochs. The LSTM converges more slowly than the parametric models.}
  \label{fig:dkt_loss}
\end{figure}

Training uses Adam with learning rate 0.001 and mini-batches of 100 students for 200 epochs. Gradients are clipped at norm 5.0. The loss is binary cross-entropy on the next-step prediction. No early stopping or learning rate schedule is applied. Hyperparameters were selected via a search over hidden dimension $\in \{64, 128, 200, 256\}$, learning rate $\in \{0.001, 0.01\}$, and dropout $\in \{0.2, 0.5\}$. Training loss curves are shown in Figure~\ref{fig:dkt_loss}.

\subsection{Code-DKT}
\label{sec:apd_codedkt}

Code-DKT extends DKT by fusing a vector representation of each submitted program with the interaction encoding at every timestep of the recurrent model. The original extracts code2vec AST-path features from Java submissions; no such extractor exists for C++, so our adaptation uses the same pretrained code embeddings that the RSSM consumes (Qwen3-Embedding-8B over the raw submission text), giving both code-aware models identical code signal. Each submission's 4096-dimensional embedding is projected to 32 dimensions by a learned linear layer with dropout, multiplied by an availability mask, and concatenated with the one-hot interaction encoding before the LSTM. All test cases of one submission share that submission's embedding.

\begin{figure}[!htbp]
  \centering
  \includegraphics[width=\columnwidth]{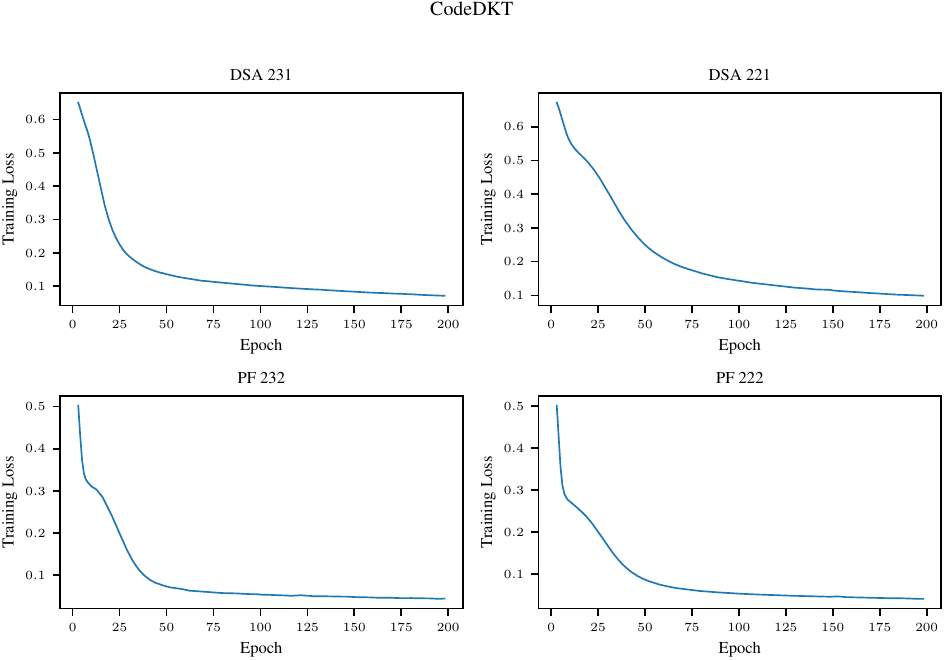}
  \caption{Code-DKT training loss across epochs for all four course offerings.}
  \label{fig:codedkt_loss}
\end{figure}

Training matches DKT's student-split protocol: Adam with learning rate 0.001 and mini-batches of 100 students for 200 epochs, hidden dimension 64, dropout 0.2, gradient clipping at norm 5.0, binary cross-entropy on the next-step prediction, and sequences truncated to the most recent 600 interactions. Hyperparameters were selected via a search over projection dimension $\in \{32, 64\}$ and dropout $\in \{0.2, 0.5\}$. Training loss curves are shown in Figure~\ref{fig:codedkt_loss}.

\subsection{TIKTOC}
\label{sec:apd_tiktoc}

TIKTOC~\citep{duan2025} combines the objectives of OKT~\citep{liu-etal-2022-open} and Code-DKT in a multi-task model: alongside generating the student's next code submission with a finetuned language model, it predicts per-test-case pass/fail outcomes from the language model's prompt hidden states. An LSTM knowledge estimator consumes the sequence of question and code embeddings, and its hidden state is injected additively into the prompt token embeddings before the language model processes them. Per-question scorer heads predict each test case's outcome from the mean-pooled prompt hidden states. The training loss is the test-case binary cross-entropy plus a code-generation cross-entropy, weighted equally in the original.

Three adaptations were required for this benchmark. First, the original ASTNN code encoder (built for Java) is replaced with the same pretrained code embeddings the RSSM and Code-DKT consume. Second, the Llama backbone is replaced with Qwen3-0.6B, finetuned with LoRA (rank 64, $\alpha = 128$, dropout 0.05) on all attention and MLP projections. Third, the original per-question prediction heads, viable at the 17 problems of the CSEDM dataset, are replaced with a single scorer shared across all questions and conditioned on the test case's text embedding and index, since with an order of magnitude more questions per-question heads see too few observations each.

\begin{figure}[!htbp]
  \centering
  \includegraphics[width=\columnwidth]{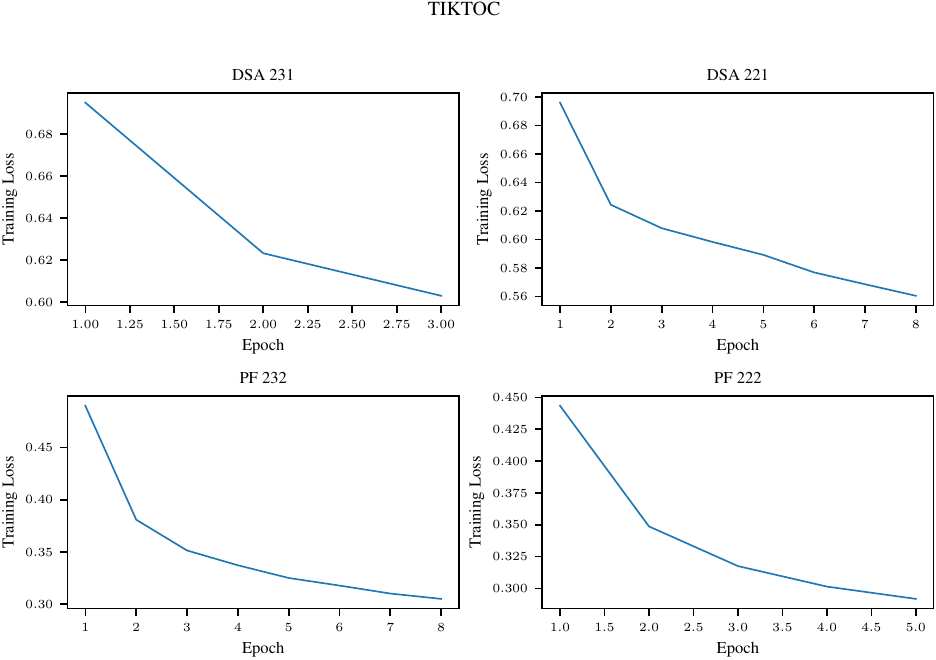}
  \caption{TIKTOC training loss across epochs for all four course offerings.}
  \label{fig:tiktoc_loss}
\end{figure}

Training uses AdamW with learning rate 1e-5 for the language model and 5e-4 for the knowledge estimator and scorer, batch size 4 with gradient accumulation of 8 (effective batch 32), up to 5 epochs with early stopping on validation test-case BCE (patience 2), gradient clipping, prompts truncated to 768 tokens and code to 512 tokens, and sequences capped at 200 interactions. Evaluation needs no code generation or execution: the scorer emits probabilities directly. Training loss curves are shown in Figure~\ref{fig:tiktoc_loss}.

\subsection{RSSM}
\label{sec:rssm_training}

The RSSM captures solver characteristics through discrete latent variables $z_t$ at each timestep, following the DreamerV2 framework. The architecture is visualized in Figure~\ref{fig:latent_rssm_apd} and consists of five components:

\begin{figure}[htbp]
    \centering
    \includegraphics[width=0.98\columnwidth]{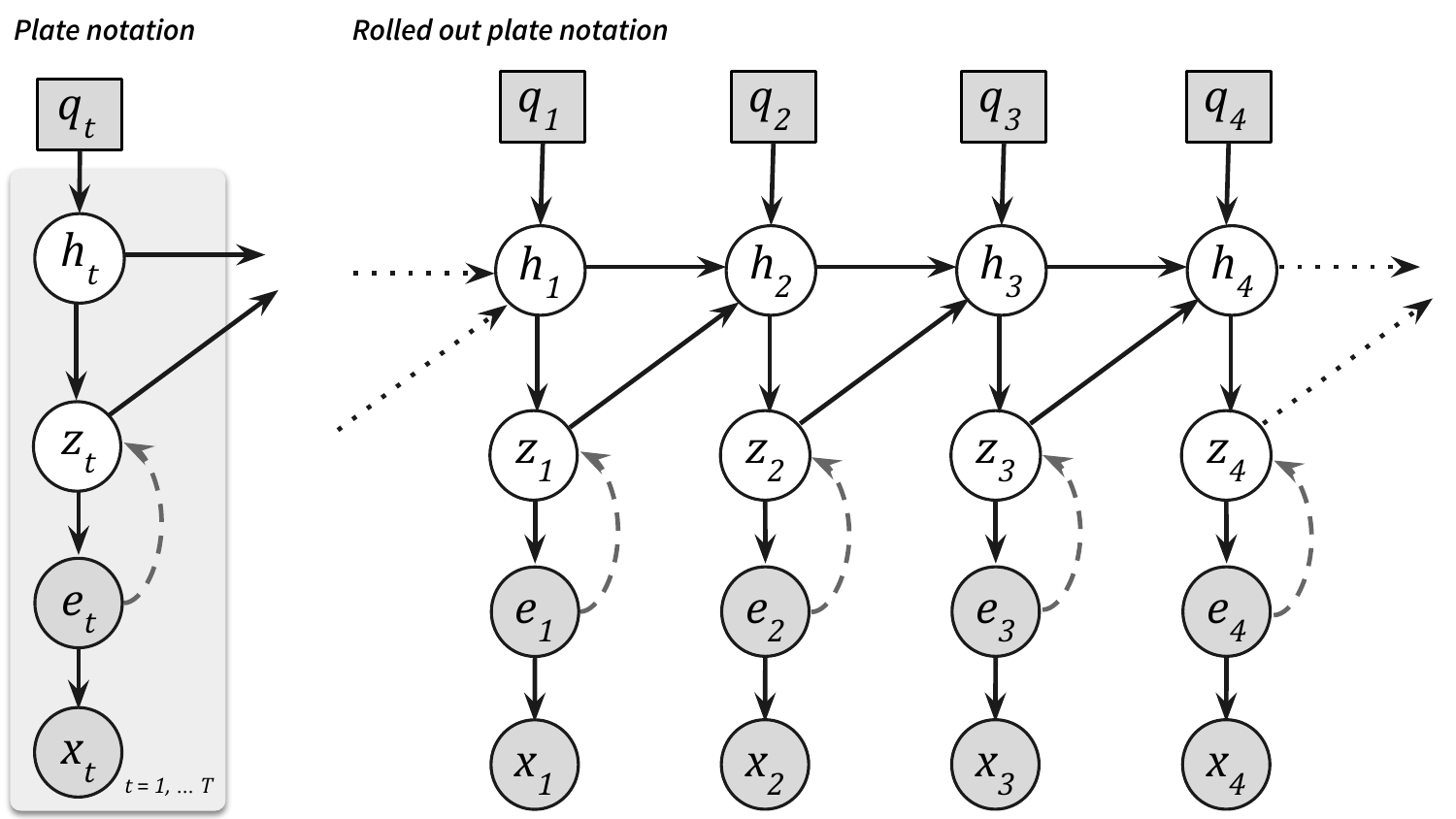}
    \caption{RSSM architecture. The GRU updates a hidden state $h_t$ from the previous discrete latent $z_{t-1}$ and problem encoding $q_t$. A posterior infers $z_t$ from $h_t$ and the response embedding $e_t$; a prior predicts $z_t$ from $h_t$ alone. The scoring head predicts per-test-case outcomes.}
    \label{fig:latent_rssm_apd}
\end{figure}

\begin{equation}
\begin{alignedat}{4}
& \text{Recurrent model:}        \quad && h_t            &\ =    &\ f_\phi(h_{t-1}, z_{t-1}, q_{t}) \\
& \text{Representation model:}   \quad && z_t            &\ \sim    &\ q_{\phi}(z_t | h_t, e_t) \\
& \text{Transition predictor:}   \quad && \hat{z}_t      &\ \sim &\ p_{\phi}(\hat{z}_t | h_t) \\
& \text{Embedding predictor:}    \quad && \hat{e}_t      &\ \sim &\ p_{\phi}(\hat{e}_t | z_t) \\
& \text{Score predictor:}        \quad && \hat{r}_t      &\ \sim &\ p_{\phi}(\hat{r}_t | q_t, h_t, z_t) \\
\end{alignedat}
\end{equation}

At each timestep $t$, the solver's submitted code is embedded into $e_t$ and the problem into $q_t$ via a pretrained LLM, each projected through learned linear layers. The GRU updates the hidden state $h_t$ from the previous latent vector $z_{t-1}$ and problem encoding $q_t$, routing information through a discrete bottleneck. The representation model (posterior) infers $z_t$ from both $h_t$ and $e_t$, while the transition predictor (prior) predicts $z_t$ from $h_t$ alone. The latent vector uses a categorical distribution with 16 variables and 16 classes, sampled via straight-through gradient estimation. The representation and transition models are MLPs with ELU activations. The score predictor takes the concatenation of $q_t$, $h_t$, and $z_t$, combined with a learned per-test-case difficulty bias.

The model uses a GRU cell with hidden dimension 512. Training uses Adam with learning rate $3 \times 10^{-4}$ and weight decay $10^{-4}$. Dropout is 0.1. Gradients are clipped at norm 1.0. The loss combines binary cross-entropy on per-test-case outcomes, MSE on embedding prediction (weight $\lambda_e{=}0.1$), and KL divergence between the posterior and prior (weight $\beta{=}0.5$) with KL balancing ($\alpha{=}0.8$), where $\mathrm{sg}$ denotes the stop-gradient operator:

\begin{equation}
\begin{aligned}
&\mathcal{L}(\phi)
\doteq
\mathbb{E}_{q_{\phi}(z_{1:T} | q_{1:T}, e_{1:T})}\Bigg[\\
    &\quad\sum_{t=1}^T \Bigg(
    \underbrace{-\ln p_{\phi}(r_t | q_t, h_t, z_t)}_{\text{score log loss}} \underbrace{+\lambda_e \lVert \hat{e}_t - e_t \rVert_2^2}_{\text{embedding loss}} \\
    &\quad+\beta\big(\alpha\, D_{KL}[\mathrm{sg}(q_{\phi}(z_t | h_t, e_t)) \| p_{\phi}(z_t | h_t)]\\
    &\quad+ (1{-}\alpha)\, D_{KL}[q_{\phi}(z_t | h_t, e_t) \| \mathrm{sg}(p_{\phi}(z_t | h_t))]\big) \Bigg)
\Bigg]
\end{aligned}
\end{equation}

\begin{figure}[!htbp]
  \centering
  \includegraphics[width=\columnwidth]{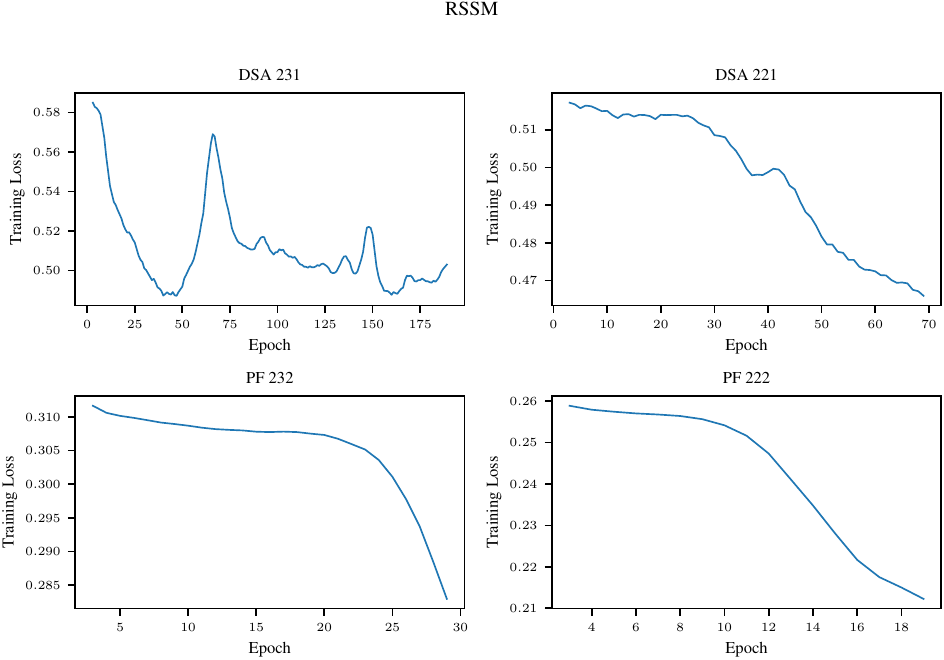}
  \caption{RSSM training loss across epochs for all four course offerings, plotted up to the checkpoint selected by validation AUC.}
  \label{fig:rssm_training_loss}
\end{figure}

Training processes students in mini-batches of 64 with Adam (learning rate $3{\times}10^{-4}$, $\beta_1{=}0.9$, $\beta_2{=}0.99$) for up to 200 epochs. The learning rate follows a cosine schedule that anneals to 10\% of its initial value by epoch 80 and is held there. Validation AUC is evaluated every 10 epochs on held-out validation students; early stopping uses patience 50, the checkpoint with the highest validation AUC is restored after training, and the test set is evaluated once on the restored checkpoint. Hyperparameters were selected via a search over hidden dimension $\in \{128, 512, 1024\}$, encoder depth, dropout $\in \{0.0, 0.1\}$, latent grid size $\in \{16{\times}16, 32{\times}32\}$, KL weight $\beta \in \{0.1, 0.5, 1.0\}$, maximum sequence length $\in \{200, 400, 600\}$. Training loss curves are shown in Figure~\ref{fig:rssm_training_loss}.

\subsection{Evaluation Data Filtering}
\label{sec:apd_filtering}

The evaluation protocol filters problems on two criteria computed over students' final submissions: the pass rate must lie in $[10\%, 90\%]$, which removes problems that nearly all students pass or fail, and more than 25\% of students must have attempted the problem. No student-level filters are applied. Table~\ref{tab:filtering_funnel} reports the students, problems, test cases, and submissions remaining after each filtering step for every course.

\input{tables/filtering_funnel}

To test whether these thresholds drive the results, we re-ran the evaluation on DSA HK231 with each filter changed in turn: the pass-rate floor, ceiling, and full band removed, all filters removed, the attempt cap lowered to 5 and raised to 15, and the coverage threshold moved to 0.15 and 0.50. Each configuration defines a different evaluation universe, so each model in Table~\ref{tab:filter_sensitivity} is retrained per configuration. The table reports test AUC. The ordering is stable: the RSSM leads in every configuration, followed by CIRT, IRT, and then BKT, with DKT and Code-DKT trailing throughout. Removing the pass-rate floor raises AUC for all models because problems that nearly all students fail are easy to discriminate; the band therefore makes the benchmark harder and does not favor any model.

\input{tables/filter_sensitivity}

\subsection{Model Efficiency}

Table~\ref{tab:efficiency} reports trainable parameter counts and measured inference time per observation for all models.

\input{tables/efficiency}

\subsection{Evaluation Metrics}

All trained models output a probability that the solver passes each test case on the next attempt. For accuracy-based metrics, predictions are binarized at a fixed threshold of 0.5. Balanced accuracy is the mean of the true-positive and true-negative rates under this threshold, which accounts for the class imbalance in pass outcomes across courses and attempt numbers. Table~\ref{tab:per_attempt_metrics} reports balanced accuracy, AUC, log loss, and Brier score for every model at each attempt number on all four courses, computed on the same student-split predictions; its balanced accuracy rows provide the full numbers underlying Figure~\ref{fig:accuracy_vs_attempt}. The LLM predictor generates code and produces binary pass/fail outcomes per test case rather than probabilities, so threshold-free metrics are not defined for it and only its balanced accuracy is reported.

\input{tables/per_attempt_metrics}

%% file: tables/filtering_funnel.tex
\begin{table}[t]
  \centering
  \small
  \caption{Data remaining after each problem-level filtering step of the evaluation protocol, per course. Pass rates are computed over students' final submissions.}
  \label{tab:filtering_funnel}
  \setlength{\tabcolsep}{3pt}
  \resizebox{\columnwidth}{!}{%
  \begin{tabular}{lrrrr}
    \toprule
    Filter step & Students & Problems & Test cases & Submissions \\
    \midrule
    \multicolumn{5}{l}{\textit{DSA HK231}} \\
    All loaded data & 965 & 227 & 2,194 & 349,929 \\
    Pass rate in [10\%, 90\%] & 965 & 157 & 1,530 & 266,861 \\
    Coverage $>$ 25\% of students & 965 & 87 & 845 & 235,453 \\
    \midrule
    \multicolumn{5}{l}{\textit{DSA HK221}} \\
    All loaded data & 507 & 175 & 1,661 & 192,988 \\
    Pass rate in [10\%, 90\%] & 507 & 119 & 1,154 & 136,921 \\
    Coverage $>$ 25\% of students & 507 & 78 & 747 & 121,004 \\
    \midrule
    \multicolumn{5}{l}{\textit{PF HK232}} \\
    All loaded data & 1,478 & 129 & 1,307 & 502,471 \\
    Pass rate in [10\%, 90\%] & 1,478 & 53 & 543 & 217,149 \\
    Coverage $>$ 25\% of students & 1,478 & 38 & 383 & 214,533 \\
    \midrule
    \multicolumn{5}{l}{\textit{PF HK222}} \\
    All loaded data & 1,399 & 119 & 1,201 & 592,269 \\
    Pass rate in [10\%, 90\%] & 1,399 & 43 & 434 & 271,271 \\
    Coverage $>$ 25\% of students & 1,399 & 37 & 375 & 269,097 \\
    \bottomrule
  \end{tabular}}
\end{table}

%% file: tables/filter_sensitivity.tex
\begin{table}[t]
  \centering
  \caption{Filter sensitivity on DSA HK231: test AUC with one filter changed at a time from the baseline. Each configuration defines a different evaluation universe of $n$ test observations, so comparisons are within rows.}
  \label{tab:filter_sensitivity}
  \resizebox{\columnwidth}{!}{%
  \begin{tabular}{lrrrrrrr}
    \toprule
    Configuration & $n$ & IRT & CIRT & BKT & DKT & CodeDKT & RSSM \\
    \midrule
    Baseline & 128,261 & 0.804 & 0.813 & 0.789 & 0.628 & 0.693 & \textbf{0.831} \\
    Pass floor off & 165,707 & 0.861 & 0.867 & 0.818 & 0.729 & 0.746 & \textbf{0.876} \\
    Pass ceiling off & 128,261 & 0.804 & 0.813 & 0.788 & 0.618 & 0.662 & \textbf{0.827} \\
    No pass band & 165,707 & 0.861 & 0.867 & 0.818 & 0.722 & 0.745 & \textbf{0.876} \\
    No filters & 186,260 & 0.853 & 0.859 & 0.806 & 0.733 & 0.731 & \textbf{0.867} \\
    Attempts 5 & 89,317 & 0.813 & 0.819 & 0.784 & 0.626 & 0.724 & \textbf{0.834} \\
    Attempts 15 & 158,841 & 0.822 & 0.828 & 0.797 & 0.696 & 0.690 & \textbf{0.843} \\
    Coverage 0.15 & 130,156 & 0.803 & 0.812 & 0.787 & 0.663 & 0.687 & \textbf{0.824} \\
    Coverage 0.50 & 126,131 & 0.806 & 0.815 & 0.792 & 0.620 & 0.702 & \textbf{0.834} \\
    \bottomrule
  \end{tabular}}
\end{table}

%% file: tables/efficiency.tex
\begin{table}[t]
  \centering
  \small
  \caption{Trainable parameters and inference time per observation. The LLM entry is generation wall clock per attempt; its parameters are frozen.}
  \label{tab:efficiency}
  \resizebox{\columnwidth}{!}{%
  \begin{tabular}{lrrl}
    \toprule
    Model & Parameters & Inference / obs. & Hardware \\
    \midrule
    IRT & 1.4K & $<$0.1 $\mu$s & Apple M5 Pro \\
    CIRT & 2.2K & $<$0.1 $\mu$s & Apple M5 Pro \\
    BKT & 3.4K & 2.0 $\mu$s & Apple M5 Pro \\
    DKT & 504.5K & 8.0 $\mu$s & Apple M5 Pro \\
    Code-DKT & 643.8K & 3.5 $\mu$s & Apple M5 Pro \\
    RSSM & 7.78M & 120.2 $\mu$s & Apple M5 Pro \\
    TIKTOC & 0.6B & 38.3 ms & Apple M5 Pro \\
    LLM (Qwen3-14B) & 14.8B (frozen) & 2.7 s & A100-40GB (vLLM) \\
    \bottomrule
  \end{tabular}}
\end{table}

%% file: tables/per_attempt_metrics.tex
\onecolumn
{\footnotesize
\renewcommand{\arraystretch}{0.92}
\setlength{\tabcolsep}{4pt}
\begin{longtable}{lllcccccccccc}
  \caption{Balanced accuracy, AUC, log loss, and Brier score at each attempt number for all models on all four courses. Balanced accuracy corresponds to Figure~\ref{fig:accuracy_vs_attempt}; predictions are binarized at 0.5. Bold marks the best model at each attempt within each course and metric (highest for balanced accuracy and AUC, lowest for log loss and Brier). The LLM predictor produces binary pass/fail outcomes rather than probabilities, so only balanced accuracy is defined for it. Entries are omitted where an attempt has fewer than 10 test observations.}
  \label{tab:per_attempt_metrics}\\
    \toprule
    Course & Model & Metric & 1 & 2 & 3 & 4 & 5 & 6 & 7 & 8 & 9 & 10 \\
    \midrule
  \endfirsthead
    \multicolumn{13}{l}{\tablename~\thetable{} (continued)}\\
    \toprule
    Course & Model & Metric & 1 & 2 & 3 & 4 & 5 & 6 & 7 & 8 & 9 & 10 \\
    \midrule
  \endhead
    \midrule
  \endfoot
    \bottomrule
  \endlastfoot
    DSA HK231 & IRT & Bal. Acc & 0.700 & 0.746 & 0.746 & 0.752 & 0.736 & 0.709 & 0.698 & 0.710 & 0.701 & 0.706 \\
     &  & AUC & 0.792 & 0.831 & 0.825 & 0.825 & 0.816 & 0.784 & 0.778 & 0.779 & 0.772 & 0.797 \\
     &  & Log loss & 0.563 & 0.513 & 0.510 & 0.509 & 0.518 & 0.544 & 0.548 & 0.550 & 0.550 & 0.521 \\
     &  & Brier & 0.191 & 0.171 & 0.170 & 0.170 & 0.174 & 0.185 & 0.187 & 0.186 & 0.187 & 0.177 \\
    \addlinespace[2pt]
     & CIRT & Bal. Acc & \textbf{0.719} & \textbf{0.755} & 0.748 & 0.752 & 0.735 & 0.705 & 0.699 & 0.706 & 0.685 & 0.688 \\
     &  & AUC & 0.799 & 0.833 & 0.823 & 0.825 & 0.819 & 0.790 & 0.791 & 0.787 & 0.782 & 0.796 \\
     &  & Log loss & \textbf{0.549} & \textbf{0.506} & 0.512 & 0.509 & 0.514 & 0.536 & 0.526 & 0.526 & 0.525 & 0.507 \\
     &  & Brier & \textbf{0.185} & \textbf{0.168} & 0.171 & 0.170 & 0.172 & 0.181 & 0.178 & 0.176 & 0.177 & 0.170 \\
    \addlinespace[2pt]
     & BKT & Bal. Acc & 0.715 & 0.745 & 0.730 & 0.739 & 0.726 & 0.710 & 0.694 & 0.702 & 0.701 & 0.711 \\
     &  & AUC & 0.776 & 0.822 & 0.808 & 0.809 & 0.793 & 0.776 & 0.762 & 0.766 & 0.762 & 0.784 \\
     &  & Log loss & 0.589 & 0.559 & 0.578 & 0.577 & 0.593 & 0.604 & 0.618 & 0.625 & 0.618 & 0.601 \\
     &  & Brier & 0.202 & 0.189 & 0.198 & 0.197 & 0.205 & 0.210 & 0.217 & 0.220 & 0.218 & 0.210 \\
    \addlinespace[2pt]
     & DKT & Bal. Acc & 0.578 & 0.585 & 0.603 & 0.607 & 0.605 & 0.581 & 0.595 & 0.586 & 0.587 & 0.588 \\
     &  & AUC & 0.610 & 0.624 & 0.641 & 0.647 & 0.640 & 0.607 & 0.623 & 0.620 & 0.628 & 0.627 \\
     &  & Log loss & 1.017 & 0.989 & 0.922 & 0.916 & 0.913 & 0.959 & 0.912 & 0.885 & 0.885 & 0.878 \\
     &  & Brier & 0.303 & 0.297 & 0.275 & 0.273 & 0.272 & 0.283 & 0.270 & 0.271 & 0.268 & 0.266 \\
    \addlinespace[2pt]
     & Code-DKT & Bal. Acc & 0.612 & 0.632 & 0.635 & 0.639 & 0.638 & 0.629 & 0.610 & 0.616 & 0.622 & 0.627 \\
     &  & AUC & 0.658 & 0.689 & 0.695 & 0.700 & 0.699 & 0.680 & 0.659 & 0.668 & 0.671 & 0.678 \\
     &  & Log loss & 0.946 & 0.884 & 0.833 & 0.826 & 0.817 & 0.850 & 0.880 & 0.858 & 0.866 & 0.844 \\
     &  & Brier & 0.287 & 0.269 & 0.255 & 0.253 & 0.250 & 0.257 & 0.267 & 0.262 & 0.259 & 0.257 \\
    \addlinespace[2pt]
     & TIKTOC & Bal. Acc & 0.647 & 0.705 & 0.707 & 0.717 & 0.694 & 0.689 & 0.683 & 0.683 & 0.673 & 0.670 \\
     &  & AUC & 0.728 & 0.784 & 0.778 & 0.796 & 0.779 & 0.775 & 0.767 & 0.759 & 0.747 & 0.757 \\
     &  & Log loss & 0.628 & 0.586 & 0.565 & 0.547 & 0.557 & 0.550 & 0.543 & 0.546 & 0.554 & 0.551 \\
     &  & Brier & 0.219 & 0.200 & 0.191 & 0.184 & 0.188 & 0.187 & 0.185 & 0.186 & 0.190 & 0.187 \\
    \addlinespace[2pt]
     & RSSM & Bal. Acc & 0.716 & 0.753 & \textbf{0.763} & \textbf{0.769} & \textbf{0.761} & \textbf{0.734} & \textbf{0.725} & \textbf{0.722} & \textbf{0.716} & \textbf{0.715} \\
     &  & AUC & \textbf{0.800} & \textbf{0.838} & \textbf{0.852} & \textbf{0.860} & \textbf{0.848} & \textbf{0.827} & \textbf{0.817} & \textbf{0.813} & \textbf{0.818} & \textbf{0.827} \\
     &  & Log loss & 0.550 & 0.509 & \textbf{0.466} & \textbf{0.462} & \textbf{0.475} & \textbf{0.491} & \textbf{0.493} & \textbf{0.496} & \textbf{0.490} & \textbf{0.475} \\
     &  & Brier & 0.186 & \textbf{0.168} & \textbf{0.156} & \textbf{0.152} & \textbf{0.157} & \textbf{0.165} & \textbf{0.167} & \textbf{0.167} & \textbf{0.165} & \textbf{0.161} \\
    \addlinespace[2pt]
     & LLM & Bal. Acc & 0.651 & 0.604 & 0.629 & 0.583 & 0.594 & 0.595 & 0.624 & 0.628 & 0.619 & 0.650 \\
    \midrule
    DSA HK221 & IRT & Bal. Acc & 0.617 & 0.669 & 0.666 & 0.683 & 0.682 & 0.680 & 0.671 & 0.658 & 0.654 & 0.639 \\
     &  & AUC & 0.736 & 0.791 & 0.781 & 0.810 & 0.801 & 0.789 & 0.795 & 0.770 & 0.770 & 0.768 \\
     &  & Log loss & 0.598 & 0.532 & 0.492 & \textbf{0.459} & 0.471 & 0.471 & 0.469 & 0.491 & 0.489 & 0.478 \\
     &  & Brier & 0.200 & 0.176 & 0.161 & \textbf{0.149} & 0.154 & 0.153 & 0.153 & 0.161 & 0.160 & 0.156 \\
    \addlinespace[2pt]
     & CIRT & Bal. Acc & \textbf{0.641} & \textbf{0.683} & \textbf{0.672} & 0.686 & 0.680 & 0.677 & 0.666 & 0.656 & 0.646 & 0.619 \\
     &  & AUC & \textbf{0.737} & 0.791 & 0.776 & 0.808 & 0.804 & 0.790 & 0.797 & 0.773 & 0.775 & 0.775 \\
     &  & Log loss & \textbf{0.578} & 0.526 & 0.497 & 0.461 & 0.468 & 0.471 & 0.466 & 0.488 & 0.483 & 0.472 \\
     &  & Brier & \textbf{0.194} & \textbf{0.174} & 0.163 & 0.150 & 0.153 & 0.152 & 0.152 & 0.159 & 0.158 & 0.154 \\
    \addlinespace[2pt]
     & BKT & Bal. Acc & 0.578 & 0.598 & 0.633 & 0.615 & 0.613 & 0.604 & 0.614 & 0.585 & 0.577 & 0.591 \\
     &  & AUC & 0.619 & 0.663 & 0.680 & 0.691 & 0.683 & 0.670 & 0.672 & 0.645 & 0.633 & 0.650 \\
     &  & Log loss & 0.679 & 0.664 & 0.666 & 0.665 & 0.660 & 0.666 & 0.665 & 0.670 & 0.676 & 0.673 \\
     &  & Brier & 0.243 & 0.236 & 0.237 & 0.236 & 0.234 & 0.237 & 0.236 & 0.239 & 0.241 & 0.240 \\
    \addlinespace[2pt]
     & DKT & Bal. Acc & 0.599 & 0.649 & 0.647 & 0.663 & 0.680 & 0.644 & 0.669 & 0.647 & 0.642 & 0.658 \\
     &  & AUC & 0.652 & 0.702 & 0.704 & 0.730 & 0.738 & 0.702 & 0.731 & 0.710 & 0.686 & 0.704 \\
     &  & Log loss & 0.812 & 0.719 & 0.659 & 0.602 & 0.594 & 0.633 & 0.595 & 0.627 & 0.666 & 0.625 \\
     &  & Brier & 0.241 & 0.217 & 0.201 & 0.189 & 0.183 & 0.195 & 0.182 & 0.193 & 0.198 & 0.185 \\
    \addlinespace[2pt]
     & Code-DKT & Bal. Acc & 0.578 & 0.609 & 0.626 & 0.634 & 0.633 & 0.621 & 0.646 & 0.658 & 0.629 & 0.640 \\
     &  & AUC & 0.630 & 0.660 & 0.677 & 0.698 & 0.693 & 0.668 & 0.701 & 0.712 & 0.681 & 0.693 \\
     &  & Log loss & 0.792 & 0.749 & 0.687 & 0.658 & 0.668 & 0.695 & 0.650 & 0.643 & 0.680 & 0.649 \\
     &  & Brier & 0.252 & 0.239 & 0.220 & 0.212 & 0.214 & 0.221 & 0.209 & 0.207 & 0.220 & 0.209 \\
    \addlinespace[2pt]
     & TIKTOC & Bal. Acc & 0.588 & 0.635 & 0.629 & 0.644 & 0.599 & 0.596 & 0.602 & 0.608 & 0.598 & 0.601 \\
     &  & AUC & 0.671 & 0.743 & 0.735 & 0.779 & 0.732 & 0.733 & 0.735 & 0.734 & 0.744 & 0.737 \\
     &  & Log loss & 0.616 & 0.564 & 0.528 & 0.491 & 0.527 & 0.514 & 0.510 & 0.516 & 0.504 & 0.487 \\
     &  & Brier & 0.212 & 0.191 & 0.176 & 0.162 & 0.176 & 0.171 & 0.170 & 0.171 & 0.167 & 0.160 \\
    \addlinespace[2pt]
     & RSSM & Bal. Acc & 0.628 & 0.682 & 0.667 & \textbf{0.696} & \textbf{0.696} & \textbf{0.683} & \textbf{0.691} & \textbf{0.662} & \textbf{0.680} & \textbf{0.678} \\
     &  & AUC & 0.728 & \textbf{0.795} & \textbf{0.785} & \textbf{0.814} & \textbf{0.819} & \textbf{0.821} & \textbf{0.821} & \textbf{0.795} & \textbf{0.802} & \textbf{0.810} \\
     &  & Log loss & 0.599 & \textbf{0.521} & \textbf{0.487} & \textbf{0.459} & \textbf{0.455} & \textbf{0.448} & \textbf{0.448} & \textbf{0.473} & \textbf{0.464} & \textbf{0.446} \\
     &  & Brier & 0.200 & \textbf{0.174} & \textbf{0.160} & \textbf{0.149} & \textbf{0.148} & \textbf{0.146} & \textbf{0.146} & \textbf{0.156} & \textbf{0.152} & \textbf{0.145} \\
    \addlinespace[2pt]
     & LLM & Bal. Acc & 0.623 & 0.591 & 0.610 & 0.636 & 0.614 & 0.580 & 0.582 & 0.570 & 0.569 & 0.579 \\
    \midrule
    PF HK232 & IRT & Bal. Acc & \textbf{0.883} & 0.836 & 0.828 & 0.804 & 0.828 & 0.830 & 0.778 & 0.825 & 0.909 & 0.839 \\
     &  & AUC & 0.935 & 0.913 & 0.907 & 0.910 & 0.888 & 0.919 & 0.839 & 0.894 & 0.942 & 0.858 \\
     &  & Log loss & 0.094 & 0.122 & 0.126 & 0.142 & 0.134 & 0.121 & 0.180 & 0.125 & 0.077 & 0.136 \\
     &  & Brier & 0.020 & 0.028 & 0.029 & 0.035 & 0.030 & 0.029 & 0.042 & 0.028 & 0.014 & 0.029 \\
    \addlinespace[2pt]
     & CIRT & Bal. Acc & \textbf{0.883} & 0.836 & 0.828 & 0.804 & 0.827 & 0.830 & 0.778 & 0.825 & 0.909 & 0.839 \\
     &  & AUC & 0.936 & 0.911 & 0.908 & 0.911 & 0.897 & 0.922 & 0.855 & 0.919 & 0.963 & 0.875 \\
     &  & Log loss & 0.091 & 0.122 & 0.126 & 0.141 & 0.132 & 0.121 & 0.176 & 0.121 & 0.083 & 0.144 \\
     &  & Brier & \textbf{0.019} & 0.028 & 0.029 & 0.035 & 0.030 & 0.029 & 0.043 & 0.028 & 0.017 & 0.032 \\
    \addlinespace[2pt]
     & BKT & Bal. Acc & 0.880 & 0.839 & 0.833 & 0.807 & 0.828 & 0.834 & 0.785 & 0.838 & 0.906 & 0.835 \\
     &  & AUC & 0.928 & 0.910 & 0.912 & 0.923 & 0.898 & 0.923 & 0.865 & 0.907 & 0.964 & 0.870 \\
     &  & Log loss & 0.211 & 0.224 & 0.223 & 0.226 & 0.229 & 0.224 & 0.252 & 0.235 & 0.218 & 0.245 \\
     &  & Brier & 0.056 & 0.059 & 0.059 & 0.061 & 0.061 & 0.059 & 0.070 & 0.062 & 0.056 & 0.066 \\
    \addlinespace[2pt]
     & DKT & Bal. Acc & 0.876 & 0.832 & 0.826 & 0.804 & 0.827 & 0.834 & 0.777 & 0.833 & 0.905 & 0.835 \\
     &  & AUC & 0.909 & 0.852 & 0.847 & 0.843 & 0.833 & 0.876 & 0.787 & 0.877 & 0.906 & 0.833 \\
     &  & Log loss & 0.130 & 0.184 & 0.187 & 0.208 & 0.198 & 0.167 & 0.261 & 0.162 & 0.108 & 0.186 \\
     &  & Brier & 0.028 & 0.037 & 0.038 & 0.044 & 0.038 & 0.035 & 0.050 & 0.034 & 0.019 & 0.035 \\
    \addlinespace[2pt]
     & Code-DKT & Bal. Acc & 0.882 & 0.834 & 0.827 & 0.802 & 0.824 & 0.827 & 0.774 & 0.820 & 0.909 & 0.836 \\
     &  & AUC & 0.911 & 0.850 & 0.845 & 0.832 & 0.827 & 0.834 & 0.800 & 0.848 & 0.894 & 0.804 \\
     &  & Log loss & 0.119 & 0.175 & 0.178 & 0.204 & 0.195 & 0.186 & 0.240 & 0.174 & 0.120 & 0.207 \\
     &  & Brier & 0.024 & 0.034 & 0.035 & 0.042 & 0.038 & 0.036 & 0.049 & 0.036 & 0.022 & 0.037 \\
    \addlinespace[2pt]
     & TIKTOC & Bal. Acc & \textbf{0.883} & \textbf{0.848} & \textbf{0.848} & \textbf{0.835} & \textbf{0.865} & \textbf{0.859} & \textbf{0.828} & \textbf{0.866} & \textbf{0.949} & \textbf{0.858} \\
     &  & AUC & 0.932 & \textbf{0.977} & \textbf{0.980} & \textbf{0.986} & \textbf{0.991} & \textbf{0.984} & \textbf{0.988} & \textbf{0.991} & \textbf{0.999} & \textbf{0.992} \\
     &  & Log loss & 0.101 & \textbf{0.076} & \textbf{0.068} & \textbf{0.078} & \textbf{0.064} & \textbf{0.071} & \textbf{0.080} & \textbf{0.064} & 0.035 & \textbf{0.064} \\
     &  & Brier & 0.021 & \textbf{0.021} & \textbf{0.020} & \textbf{0.024} & \textbf{0.020} & \textbf{0.022} & \textbf{0.026} & \textbf{0.020} & 0.010 & \textbf{0.019} \\
    \addlinespace[2pt]
     & RSSM & Bal. Acc & \textbf{0.883} & 0.836 & 0.828 & 0.806 & 0.829 & 0.830 & 0.778 & 0.825 & 0.909 & 0.839 \\
     &  & AUC & \textbf{0.939} & 0.952 & 0.974 & 0.972 & 0.973 & 0.980 & 0.958 & 0.977 & \textbf{0.999} & 0.927 \\
     &  & Log loss & \textbf{0.088} & 0.108 & 0.099 & 0.112 & 0.099 & 0.090 & 0.131 & 0.085 & \textbf{0.033} & 0.110 \\
     &  & Brier & \textbf{0.019} & 0.026 & 0.026 & 0.031 & 0.026 & 0.024 & 0.035 & 0.022 & \textbf{0.007} & 0.024 \\
    \midrule
    PF HK222 & IRT & Bal. Acc & 0.712 & 0.746 & 0.762 & 0.756 & 0.742 & 0.754 & 0.761 & 0.769 & 0.752 & 0.765 \\
     &  & AUC & 0.859 & 0.880 & 0.889 & 0.881 & 0.881 & 0.892 & 0.888 & 0.885 & 0.886 & 0.893 \\
     &  & Log loss & 0.323 & 0.278 & 0.261 & 0.266 & 0.281 & 0.264 & 0.258 & 0.256 & 0.259 & 0.245 \\
     &  & Brier & 0.100 & 0.083 & 0.077 & 0.078 & 0.084 & 0.078 & 0.075 & 0.074 & 0.076 & 0.072 \\
    \addlinespace[2pt]
     & CIRT & Bal. Acc & 0.711 & 0.743 & 0.759 & 0.756 & 0.744 & 0.753 & 0.762 & 0.775 & 0.761 & 0.776 \\
     &  & AUC & 0.862 & 0.880 & 0.889 & 0.880 & 0.882 & 0.894 & 0.888 & 0.890 & 0.893 & 0.897 \\
     &  & Log loss & 0.321 & 0.278 & 0.261 & 0.266 & 0.280 & 0.262 & 0.257 & 0.252 & 0.253 & 0.242 \\
     &  & Brier & 0.099 & 0.083 & 0.077 & 0.078 & 0.083 & 0.078 & 0.075 & 0.073 & 0.074 & 0.071 \\
    \addlinespace[2pt]
     & BKT & Bal. Acc & 0.659 & 0.693 & 0.709 & 0.716 & 0.699 & 0.717 & 0.731 & 0.740 & 0.733 & 0.747 \\
     &  & AUC & 0.815 & 0.851 & 0.866 & 0.862 & 0.860 & 0.869 & 0.878 & 0.873 & 0.872 & 0.867 \\
     &  & Log loss & 0.349 & 0.313 & 0.299 & 0.297 & 0.311 & 0.299 & 0.287 & 0.285 & 0.289 & 0.282 \\
     &  & Brier & 0.108 & 0.093 & 0.087 & 0.087 & 0.093 & 0.088 & 0.083 & 0.082 & 0.084 & 0.081 \\
    \addlinespace[2pt]
     & DKT & Bal. Acc & 0.680 & 0.725 & 0.734 & 0.742 & 0.735 & 0.744 & 0.753 & 0.760 & 0.763 & 0.771 \\
     &  & AUC & 0.793 & 0.829 & 0.834 & 0.842 & 0.838 & 0.845 & 0.839 & 0.851 & 0.867 & 0.863 \\
     &  & Log loss & 0.489 & 0.397 & 0.387 & 0.373 & 0.395 & 0.374 & 0.378 & 0.359 & 0.346 & 0.339 \\
     &  & Brier & 0.128 & 0.105 & 0.103 & 0.099 & 0.104 & 0.099 & 0.099 & 0.096 & 0.095 & 0.091 \\
    \addlinespace[2pt]
     & Code-DKT & Bal. Acc & 0.684 & 0.720 & 0.728 & 0.734 & 0.722 & 0.724 & 0.746 & 0.748 & 0.748 & 0.759 \\
     &  & AUC & 0.793 & 0.823 & 0.832 & 0.821 & 0.815 & 0.819 & 0.828 & 0.833 & 0.851 & 0.858 \\
     &  & Log loss & 0.471 & 0.383 & 0.363 & 0.371 & 0.400 & 0.383 & 0.360 & 0.354 & 0.338 & 0.316 \\
     &  & Brier & 0.121 & 0.099 & 0.095 & 0.094 & 0.102 & 0.099 & 0.092 & 0.089 & 0.090 & 0.085 \\
    \addlinespace[2pt]
     & TIKTOC & Bal. Acc & 0.669 & 0.735 & 0.745 & 0.760 & 0.756 & 0.758 & 0.766 & 0.777 & 0.778 & 0.780 \\
     &  & AUC & 0.864 & \textbf{0.936} & 0.936 & 0.936 & 0.941 & 0.934 & 0.928 & 0.934 & 0.933 & 0.945 \\
     &  & Log loss & 0.330 & 0.225 & 0.217 & 0.214 & 0.223 & 0.225 & 0.223 & 0.215 & 0.219 & 0.199 \\
     &  & Brier & 0.101 & 0.070 & 0.067 & 0.065 & 0.069 & 0.068 & 0.066 & 0.064 & 0.065 & 0.061 \\
    \addlinespace[2pt]
     & RSSM & Bal. Acc & \textbf{0.719} & \textbf{0.757} & \textbf{0.768} & \textbf{0.772} & \textbf{0.763} & \textbf{0.769} & \textbf{0.777} & \textbf{0.790} & \textbf{0.794} & \textbf{0.812} \\
     &  & AUC & \textbf{0.882} & 0.935 & \textbf{0.941} & \textbf{0.942} & \textbf{0.945} & \textbf{0.938} & \textbf{0.931} & \textbf{0.943} & \textbf{0.942} & \textbf{0.952} \\
     &  & Log loss & \textbf{0.302} & \textbf{0.224} & \textbf{0.210} & \textbf{0.206} & \textbf{0.219} & \textbf{0.217} & \textbf{0.215} & \textbf{0.199} & \textbf{0.207} & \textbf{0.185} \\
     &  & Brier & \textbf{0.094} & \textbf{0.068} & \textbf{0.064} & \textbf{0.062} & \textbf{0.067} & \textbf{0.066} & \textbf{0.064} & \textbf{0.059} & \textbf{0.062} & \textbf{0.054} \\
\end{longtable}
}
\twocolumn

%% file: sections/8.9A9.LLMPredictorDetails.tex
\section{LLM-as-Predictor Details}
\label{sec:llm_predictor_appendix}

This appendix provides implementation details for the LLM-based predictor. All models share the same data split, filtering, and evaluation protocol described in Section~\ref{sec:data_setup}.

\subsection{Student Split and Temporal Boundary}

The LLM predictor uses the same train/test student split as the trained models (Section~\ref{sec:data_setup}). Students are partitioned into train and test sets. The retrieval index contains all of each train student's submissions across all weeks, plus each test student's submissions from weeks 1 through the calibration cutoff only. This ensures the LLM has access to the same temporal scope of data as the parametric baselines.

\subsection{Persona Construction}

For each test student $i$, we construct a structured behavioral persona $s_i$ from their calibration-period submissions (weeks 1--3). The persona captures: average time between submissions, average Levenshtein edit distance per attempt, total submission count, precheck-to-submit ratio, per-topic pass rates, code length and line count, and an improvement trend (improving, stable, or declining based on recent vs.\ earlier pass rates).

Students are additionally classified into one of four behavioral archetypes based on their position in the joint distribution of submission timing, edit size, and volume: rapid iterators (frequent small changes), deliberate planners (moderate pace, large rewrites), careful thinkers (long pauses, few attempts), and balanced (mixed strategy). The archetype label and a natural-language description are included in the system prompt.

\subsection{Retrieval and Summarization}

From the student's calibration history, we retrieve $K{=}5$ prior questions as few-shot context using a score that combines content similarity with temporal recency:
\begin{equation}
\begin{split}
    \text{score}(q_k) ={}& (1 - \lambda)\,\text{sim}(q_k, q_t) \\
    &+ \lambda\,\exp\!\left(-\frac{\ln 2 \cdot \Delta t_k}{\tau}\right)
\end{split}
\end{equation}
where $\text{sim}(q_k, q_t)$ is the TF-IDF cosine similarity between the candidate and target question texts, $\Delta t_k$ is the time elapsed since the student last attempted $q_k$, $\tau = 4$ weeks is the decay half-life, and $\lambda = 0.5$. Retrieval is restricted to questions from weeks strictly before the target question's week, respecting the temporal boundary.

For each retrieved question, all of the student's attempts are included. To fit within context limits, each trajectory is condensed into a behavioral summary by a smaller LLM (Claude Haiku), describing the student's problem-solving strategy and common mistakes. The target question is also summarized into a one-sentence description. All summaries are pre-computed in batch before the evaluation loop begins.

\subsection{Prompt Structure}

Each prediction uses a two-part prompt. The system message contains: a task framing instruction (predict what this student would submit), the target question summary and item-level statistics (pass rate, average attempts to solve, number of students who attempted), computed from training-set students only, and the student's persona profile. The user message contains: summaries of the student's prior approaches on similar problems, the submission format instructions (precheck or submit), the full question text and code template, and (on subsequent attempts) the student's accumulating attempt trajectory.

\subsection{Prediction Protocol}

For each student-question pair, the LLM produces one prediction per attempt in the student's real trajectory. At attempt $n$, the LLM generates code conditioned on the student's trajectory through attempt $n{-}1$:
\begin{equation}
    \hat{x}_t^{(n)},\, a_t^{(n)} \sim p_{\text{LLM}}\!\left(\cdot \mid s_i,\, \tilde{\mathcal{H}}_i,\, q_t,\, w_{q_t},\, \mathcal{T}_t^{(n-1)}\right)
\end{equation}
where $\hat{x}_t^{(n)}$ is the generated code, $a_t^{(n)} \in \{\text{precheck}, \text{submit}\}$ is the chosen action, $q_t$ is the target question, and $\mathcal{T}_t^{(n-1)}$ is the student's real attempt trajectory through attempt $n{-}1$. On the first attempt, $\mathcal{T}_t^{(0)}$ is empty and the LLM sees only the prompt described above. On each subsequent attempt, the context grows to include the student's prior attempts, each showing the action type (precheck or submit) and pass pattern (e.g., \texttt{1010000000}), along with Haiku-generated summaries of the student's code. The LLM does not receive test-case details (inputs, expected outputs, or actual outputs); it sees only the binary pass pattern. The number of predictions per question matches the student's real attempt count, up to a maximum of 10. The generated code is compiled and graded: prechecks are evaluated against public test cases only, while submissions are graded against the full test suite.

We evaluate three predictor models, all with frozen parameters and no fine-tuning: Qwen3-14B and Gemma-4-31B, served with vLLM on A100 GPUs, and Claude Opus 4.6, accessed through the Anthropic API. Opus calls use standard inference mode without extended thinking, so the model generates responses directly from the prompt without an explicit chain-of-thought reasoning budget; requests are issued with up to 10 concurrent workers and a 0.2-second inter-request delay to stay within rate limits. Items are processed in chunks of 50, with results saved after each chunk to support resumption.

\subsection{Example Prompt}
\label{sec:example_prompt}

On the first attempt, the LLM receives the full system and user messages described above: persona, peer and self summaries, question text, and code template. On each subsequent attempt, the prompt accumulates Haiku-generated summaries of the student's real prior attempts, and the LLM predicts what the student would submit next. For example, on Graph BFS (Q139), the LLM's predicted submissions across one student's 4 attempts score pass patterns \texttt{1010000000} $\rightarrow$ \texttt{1010000000} $\rightarrow$ \texttt{1010000000} $\rightarrow$ \texttt{1111111111}. At attempt 3, the LLM sees summaries of the student's three prior real submissions before generating its prediction.

The first listing below shows the full initial prompt for BST Single Children, illustrating the system and user message structure. The system message provides the task framing, question metadata, and student persona. The user message provides peer and self approach summaries, the question text, the code template, and submission format instructions.

On subsequent attempts, the prompt accumulates Haiku-generated summaries of the student's real prior submissions. The second listing below shows this feedback loop for a different student on Graph BFS. At each attempt, the LLM receives the growing trajectory of prior attempt summaries and predicts the next submission. In this example, the LLM produces three identical submissions scoring 2/10, then on attempt 4, after seeing the full trajectory including the student's successful third submission, generates code that passes all 10 tests.

\paragraph{Example: Initial Prompt with Student Context.}
The listing below shows the full initial prompt (attempt~1) for BST Single Children. The system message (lines 1--24) provides the task framing, question metadata, and student behavioral persona. The user message (lines 26--end) provides peer approach summaries, the student's own prior approaches on similar problems, the question text, the C++ code template, and submission format instructions.

\begin{lstlisting}[frame=single,framesep=2mm,caption={}]
--- SYSTEM MESSAGE ---

You are predicting what a specific university student would submit for a C++ programming assignment. Your code should authentically reflect this student's ability level, coding style, and typical mistake patterns. Study the student profile and peer evidence to calibrate your response.

=== QUESTION: BST - Single Children ===

The student must implement a method to identify and return all nodes in a binary search tree that have exactly one child (either left or right, but not both), using tree traversal on the BST data structure.
Topic: Exam 78% of students pass all tests on this problem. Average 2.4 attempts to solve. 327 students attempted this question.

=== STUDENT IDENTITY ===

Background: a Data Structures and Algorithms student in a Regular section who has attempted 78 distinct problems with 325 total submissions.

Behavioral Style: Spends significant time between submissions. Makes fewer total attempts. Plans solutions carefully before coding.
- Average time between submissions: 15.0 hours
- Average code change per attempt: 192 characters (Levenshtein)
- Precheck vs Submit ratio: 31% Precheck, 69% Submit
- Tendency: Rewrites large portions of code

Overall pass rate: 18%
Topic proficiency: Sorting (Easy): 67%, Queue: 31%, Array List: 29%, Exam: 29%, Singly Linked List: 28%, Sorting (Advance): 24%, Binary Search Tree: 19%, Binary Tree: 19% (and 5 more topics)
Typical code length: 732 characters (10 lines), concise style. Attempts per problem: 4.2 on average.

Embody this student fully. Your coding ability, problem-solving approach, and submission behavior should be consistent with the profile above.

--- USER MESSAGE ---

=== How Other Students at a Similar Level Approached This Problem ===

Student 1: The student took 3 attempts to solve this BST problem, using a recursive helper function strategy to count nodes with exactly one child. Their initial attempt had a function name mismatch (calling "Recursive" instead of "RecursiveChild"), which they corrected in attempt 2. The logic itself was sound--identifying nodes with only a left or only a right child, incrementing the counter, and recursively continuing traversal--and they eventually solved it successfully on their final submission.

Student 2: The student took 9 attempts to solve this problem, struggling primarily with syntax errors (inconsistent pointer notation like `pLeft` vs `left`, and typos like `single Chile`) in the early attempts. After fixing the syntax issues by attempt 4, they then made a logical error in attempt 5 by removing the count of the current node in the base case, which they partially corrected in attempt 8 but the fundamental algorithm remained flawed--they were still counting nodes with both children in the final return statement. Despite submitting twice with what appeared to be incomplete solutions (marked "1111111111"), they eventually solved it, though the trajectory suggests they may have arrived at the correct approach through debugging or realization that nodes with two children should not be counted toward the single-child total.

Student 3: The student took 6 attempts to solve this problem, initially using an iterative stack-based in-order traversal approach that failed all test cases. After four unsuccessful attempts with the same iterative strategy (possibly with minor debugging attempts), they pivoted to a recursive approach in attempt 5, which successfully passed all test cases. The key insight was recognizing that a recursive solution checking each node's children and recursively counting single-child nodes throughout the tree was the correct strategy, whereas the iterative in-order traversal logic was fundamentally flawed for this problem.

Student 4: The student solved the problem in a single attempt. Their approach was to recursively traverse the entire binary search tree, checking each node to determine if it had exactly one child (either left or right, but not both), and accumulating a count of such nodes. The solution was correct on the first try, successfully identifying and counting all nodes with a single child through a straightforward recursive traversal.

Student 5: The student took 4 attempts to solve this problem, using a recursive approach that counts nodes with exactly one child by checking if each node has only a left or right subtree and recursing accordingly. They made a critical logic error by continuing to recurse down single-child branches instead of counting that node and stopping, which caused them to count every node in unbalanced paths rather than just the single-child nodes themselves. Despite submitting the same flawed code twice (attempts 2 and 4), they eventually corrected their approach and successfully solved the problem.

Based on this student's ability and the peer evidence above, produce code consistent with how a student at this level would approach this problem.

=== How This Student Approached Similar Problems ===

1. [BST - Find k-th smallest element] The student must implement a function to find the k-th smallest element in a Binary Search Tree using in-order traversal, which naturally visits nodes in sorted order.
Student approach: This student uses a straightforward **collect-all-then-index approach** (in-order traversal into a vector, then return k-th element), which is algorithmically correct but memory-inefficient. Their iteration strategy shows minimal actual changes--attempts 1-3 are nearly identical, suggesting they made only cosmetic adjustments (storing size in a variable) rather than debugging a real problem, indicating they may not have understood why their prechecks failed or simply resubmitted the same logic hoping it would pass.

2. [BST - Alternative Level Traversal] The student must implement a function to traverse a binary search tree and return nodes alternating between left-to-right and right-to-left at each level, using a level-order traversal approach.
Student approach: **Summary:**

This student demonstrates a **trial-and-error debugging approach with incremental refinement**: their first four attempts struggled with type mismatches (mixing `BSTNode`, `TreeNode`, and `BTNode`), then they pivoted to implementing the core algorithm logic (alternating level traversal) starting in attempt 5. They systematically tested different reversal techniques--using a stack (attempt 5), vector insertion (attempt 6), and finally index-based placement (attempt 7)--before converging on a working solution that correctly reverses alternate levels by pre-allocating a level vector and writing values at reversed indices.

3. [BST - Subtree in range] The student must implement a function to count or find all nodes in a Binary Search Tree whose values fall within a specified range [min, max] using BST properties to efficiently prune branches outside the range.
Student approach: This student demonstrates a **post-order recursive traversal strategy** with immediate node deletion when out-of-range values are encountered. However, they fail to iterate or debug between attempts--submitting identical code twice despite the first attempt failing precheck--suggesting they either didn't understand the error feedback or lacked a systematic debugging approach to identify why their logic (recursively trimming invalid branches) wasn't producing correct results.

4. [BST - Range Count] The student must implement a function that counts the number of nodes in a Binary Search Tree whose values fall within a given range [low, high] using recursive tree traversal.
Student approach: This student demonstrates a **solid algorithmic understanding** of BST range queries with a correct recursive strategy that prunes branches based on BST properties (exploring right subtree when value is too low, left subtree when too high). However, they show a **weak debugging methodology**: they submitted identical code twice without investigating why the precheck failed, suggesting they either misread the feedback or lacked a systematic approach to understanding what went wrong rather than iterating on the logic itself.

5. [BT - Lowest Ancestor] The student must implement a function to find the lowest common ancestor (LCA) of two nodes in a binary tree using tree traversal.
Student approach: **Summary:**

This student uses a systematic debugging approach, methodically fixing one issue at a time--correcting parameter mismatches (n1/n2 -> a/b), then struct field names (key -> val)--but fails to address the core algorithmic flaw: the path comparison loop doesn't track actual path lengths, causing it to compare uninitialized array values and return incorrect ancestors. Their iterative strategy of fixing syntax/naming errors masks the need for a fundamental logic redesign.

=== Submission Format ===
You can either [Precheck] (run public tests only, no grade penalty) or [Submit] (final submission, graded against all tests).
Example:
[Precheck]
```cpp
// your implementation here
```

=== Question ===
Class `BSTNode` is used to store a node in binary search tree, described on
the following:

    class BSTNode {
    public: 
        int val; 
        BSTNode *left; 
        BSTNode *right; 
        BSTNode() {
            this->left = this->right = nullptr;
        } 
        BSTNode(int val) {
            this->val = val;
            this->left = this->right = nullptr;
        } 
        BSTNode(int val, BSTNode*& left, BSTNode*& right) {
            this->val = val;
            this->left = left;
            this->right = right;
        } 
    };

Where `val` is the value of node, `left` and `right` are the pointers to the
left node and right node of it, respectively. If a repeated value is inserted
to the tree, it will be inserted to the left subtree.

Also, a static method named `createBSTree` is used to create the binary search
tree, by iterating the argument array left-to-right and repeatedly calling
`addNode` method on the root node to insert the value into the correct
position. For example:

    int arr[] = {0, 10, 20, 30};
    auto root = BSTNode::createBSTree(arr, arr + 4);

is equivalent to

    auto root = new BSTNode(0);
    root->addNode(10);
    root->addNode(20);
    root->addNode(30);

**Request:** Implement function:

` int singleChild(BSTNode* root); `

Where `root` is the root node of given binary search tree (this tree has
between 0 and 100000 elements). This function returns the number of single
children in the tree.

**More information:**

\- A node is called a **single child** if its parent has only one child.

**Example:**

Given a binary search tree in the following:

![singleChild](https://e-learning.hcmut.edu.vn/draftfile.php/177981/user/draft/713969830/singleChild.png)

There are `2` single children: node `2` and node `3`.

_Note: In this exercise, the libraries`iostream` and `using namespace std` are
used. You can write helper functions; however, you are not allowed to use
other libraries._

=== Code Template ===
#include <iostream>
using namespace std;
#define SEPARATOR "#<ab@17943918#@>#"

class BSTNode {
    private:
        void addNode(int val);
    public: 
        int val;
        BSTNode *left;
        BSTNode *right;
        BSTNode() {
            this->left = this->right = nullptr;
        };
        BSTNode(int val) {
            this->val = val;
            this->left = this->right = nullptr;
        }
        BSTNode(int val, BSTNode*& left, BSTNode*& right) {
            this->val = val;
            this->left = left;
            this->right = right;
        }
        static void printPreorder(BSTNode* root);
        static void deleteTree(BSTNode* root);
        template <typename InputIterator>
        static BSTNode* createBSTree(InputIterator first, InputIterator last);
};

void BSTNode::addNode(int val) {
    BSTNode* trav = this;
    while ((trav->left != nullptr && val <= trav->val) || (trav->right != nullptr && val > trav->val)) {
        trav = val <= trav->val ? trav->left : trav->right;
    }
    if (val <= trav->val) trav->left = new BSTNode(val);
    else trav->right = new BSTNode(val);
}

template <typename InputIterator>
BSTNode* BSTNode::createBSTree(InputIterator first, InputIterator last) {
    BSTNode* root = nullptr;
    for (; first != last; first++) {
        if (root == nullptr) root = new BSTNode(*first);
        else root->addNode(*first);
    }
    return root;
}

void BSTNode::deleteTree(BSTNode* root) {
    if (!root) return;
    deleteTree(root->left);
    deleteTree(root->right);
    delete root;
}

void BSTNode::printPreorder(BSTNode* root) {
    if (!root) return;
    cout << root->val << " ";
    printPreorder(root->left);
    printPreorder(root->right);
}

{{ STUDENT_ANSWER }}

int main() {
{% for TEST in TESTCASES %}
   {
    {{ TEST.extra }};
    {{ TEST.testcode }};
   }
    {% if not loop.last %}cout << SEPARATOR << endl;{% endif %}
{% endfor %}
    return 0;
}

You must first choose an action -- either [Precheck] (run public tests only, no grade penalty) or [Submit] (final submission, graded against all tests).

Then provide your C++ implementation that will replace the {{ STUDENT_ANSWER }} block in the template.
- Do NOT reproduce any part of the template.
- Do NOT emit `int main()` (it's already declared).
- Include any needed class definitions.

IMPORTANT: Your response format must be exactly:
1. First line: either [Precheck] or [Submit]
2. Second line: ```cpp
3. Your code
4. Last line: ```
No extra characters, whitespace, or text may appear before or after.
\end{lstlisting}

\paragraph{Example: Feedback Loop Across Attempts.}
The listing below shows the feedback loop across 4 attempts for Graph BFS. Attempt~1 uses the full initial prompt (not shown). On each subsequent attempt, the LLM receives Haiku-generated summaries of the student's real prior submissions and predicts the next submission. The trajectory accumulates: attempt~2 sees one prior summary, attempt~3 sees two, and attempt~4 sees three (including the student's successful submission with pass pattern \texttt{1111111111}). The LLM's pass pattern progresses from \texttt{1010000000} to \texttt{1111111111}.

\begin{lstlisting}[frame=single,framesep=2mm,caption={}]
========== ATTEMPT 1 ==========
[Submit]  pass=1010000000 (2/10 tests passed)

========== ATTEMPT 2 ==========

--- PROMPT ---
=== Student's Attempt Trajectory ===

Attempt 1 [Precheck] ->
Summary: The student implemented a Graph class with a BFS
(Breadth-First Search) method that traverses a graph starting from a
given vertex and returns an Adjacency object containing the order in
which vertices were visited. The implementation uses a queue to process
vertices level-by-level and a visited array to avoid revisiting
vertices.

Now predict what the student would submit next. Match their coding
style and debugging approach.

--- RESPONSE ---
[Submit]  pass=1010000000 (2/10 tests passed)

========== ATTEMPT 3 ==========

--- PROMPT ---
=== Student's Attempt Trajectory ===

Attempt 1 [Precheck] ->
Summary: The student implemented a Graph class with a BFS
(Breadth-First Search) method that traverses a graph starting from a
given vertex and returns an Adjacency object containing the order in
which vertices were visited. The implementation uses a queue to process
vertices level-by-level and a visited array to avoid revisiting
vertices.

Attempt 2 [Precheck] ->
Summary: The student implemented a Graph class with BFS
(Breadth-First Search) functionality that stores an adjacency list
representation using a custom Adjacency data structure. The BFS method
performs a traversal starting from a given vertex, marking visited
vertices, using a queue to explore neighbors level-by-level, and
returns the traversal order in a new Adjacency object.

Now predict what the student would submit next. Match their coding
style and debugging approach.

--- RESPONSE ---
[Submit]  pass=1010000000 (2/10 tests passed)

========== ATTEMPT 4 ==========

--- PROMPT ---
=== Student's Attempt Trajectory ===

Attempt 1 [Precheck] ->
Summary: The student implemented a Graph class with a BFS
(Breadth-First Search) method that traverses a graph starting from a
given vertex and returns an Adjacency object containing the order in
which vertices were visited. The implementation uses a queue to process
vertices level-by-level and a visited array to avoid revisiting
vertices.

Attempt 2 [Precheck] ->
Summary: The student implemented a Graph class with BFS
(Breadth-First Search) functionality that stores an adjacency list
representation using a custom Adjacency data structure. The BFS method
performs a traversal starting from a given vertex, marking visited
vertices, using a queue to explore neighbors level-by-level, and
returns the traversal order in a new Adjacency object.

Attempt 3 [Submit] -> 1111111111
Summary: The student implemented a Graph class with a BFS
(Breadth-First Search) method that traverses a graph starting from a
given vertex and returns an Adjacency list containing the vertices in
the order they were visited. The implementation uses a queue to process
vertices level-by-level and a visited array to avoid revisiting
vertices.

Now predict what the student would submit next. Match their coding
style and debugging approach.

--- RESPONSE ---
[Submit]  pass=1111111111 (10/10 tests passed)
\end{lstlisting}

\subsection{Prompt Ablation}
\label{sec:prompt_ablation}

We ablate the LLM predictor's prompt to isolate the contribution of each component. Table~\ref{tab:prompt_ablation} summarizes the six conditions. The \textbf{full} pipeline includes the student persona, prior attempt trajectory, and RAG-retrieved examples of the student's approach on the $K{=}5$ most similar problems. Each ablated variant removes one component while keeping the rest. The \textbf{recent questions} condition keeps all three components but selects the $K{=}5$ problems by recency of the student's last submission instead of content similarity. The \textbf{direct solve} condition removes all student-specific information: the LLM receives only the question and code template, solving each problem from scratch with the same iterative feedback loop.

\begin{table}[t]
  \centering
  \small
  \caption{Prompt ablation conditions. \cmark\ = included, \xmark\ = removed.}
  \label{tab:prompt_ablation}
  \setlength{\tabcolsep}{4pt}
  \begin{tabular}{lccc}
    \toprule
    Condition & Persona & Trajectory & RAG \\
    \midrule
    Full & \cmark & \cmark & \cmark \\
    No persona & \xmark & \cmark & \cmark \\
    No trajectory & \cmark & \xmark & \cmark \\
    No RAG & \cmark & \cmark & \xmark \\
    Recent questions & \cmark & \cmark & recency \\
    Direct solve & \xmark & \xmark & \xmark \\
    \bottomrule
  \end{tabular}
\end{table}

\begin{figure}[t]
  \centering
  \includegraphics[width=\columnwidth]{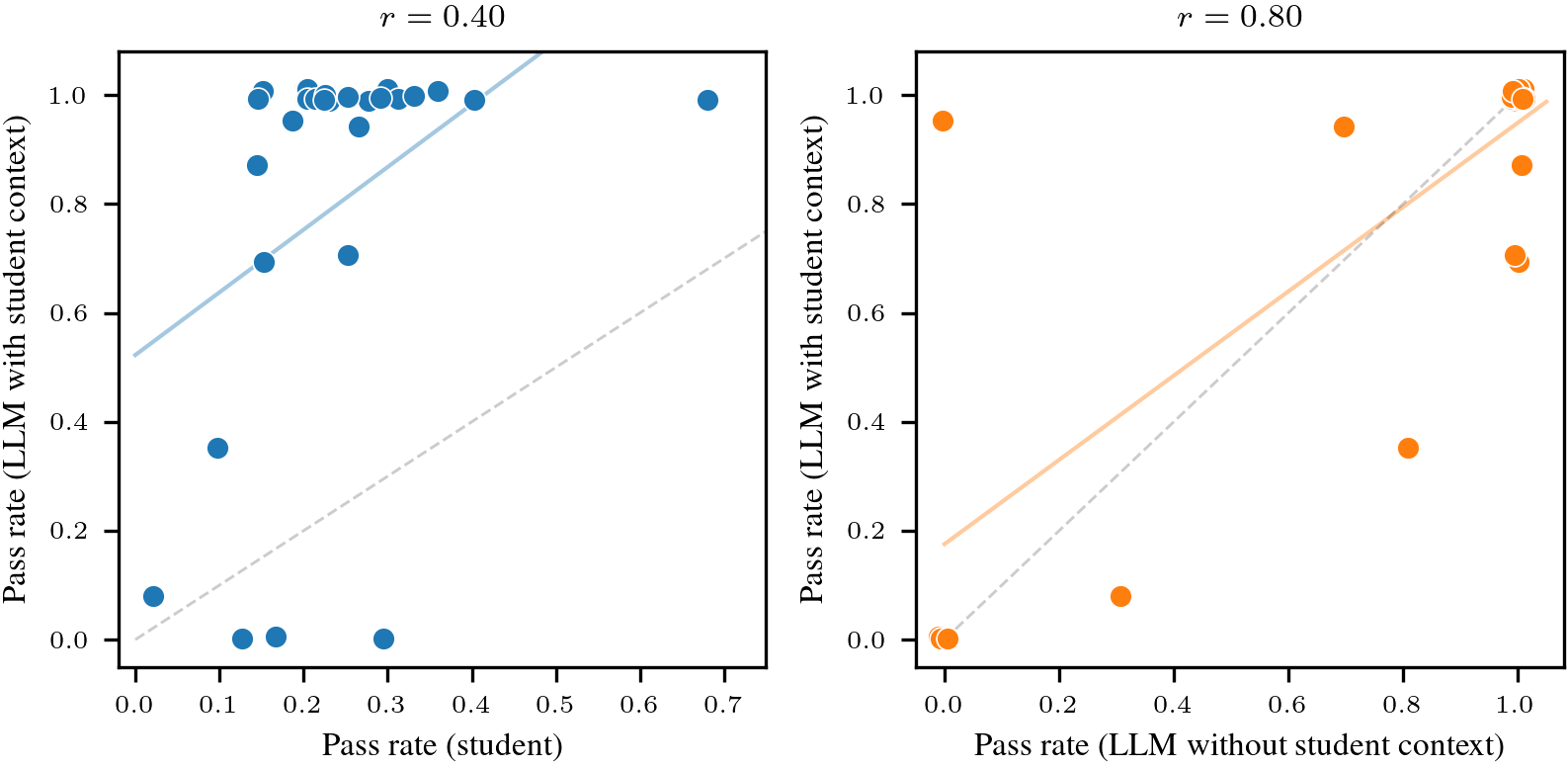}
  \caption{Problem-level pass rate correlations for Opus 4.6 on DSA HK231. Left:
  student pass rate vs.\ LLM pass rate with student context
  ($r = 0.40$). Right: LLM pass rate without student context
  vs.\ LLM pass rate with student context ($r = 0.80$). The
  stronger correlation between the two LLM conditions suggests
  the LLM's performance on each problem is driven primarily
  by its own coding ability rather than the student profile.}
  \label{fig:grounded_correlation}
\end{figure}

Results for Qwen3-14B are shown in Figure~\ref{fig:llm_ablation} (bottom panel, main text). The full pipeline achieves the highest balanced accuracy. Removing the persona produces the largest drop, with removing the trajectory context close behind, so the student profile and the attempt history together carry most of the conditioning signal. This is compatible with the difficulty-driven pattern in Figure~\ref{fig:grounded_correlation}: the persona includes the student's aggregate pass-rate statistics, which can calibrate how often predicted submissions fail without requiring student-specific dynamics. Removing RAG examples produces a smaller drop, and selecting the retrieved problems by recency instead of similarity performs comparably to removing them entirely.

Figure~\ref{fig:grounded_correlation} compares problem-level pass rates between the full and direct-solve conditions. The correlation between the LLM's pass rate without student context and its pass rate with student context is high ($r = 0.80$), while the correlation with real student pass rates is lower ($r = 0.40$). This indicates that the LLM's problem-level performance is driven primarily by its own coding ability rather than the student profile.

%% file: sections/8.1A1.UseofGenerativeAITools.tex
\clearpage
\section{Use of Generative AI Tools}

The paper has benefited from language refinement and code debugging assistance using generative AI tools, including OpenAI’s GPT family and Anthropic’s Claude 3.5. These tools were used under the authors' supervision to improve clarity, phrasing and to identify issues during the development of the work. All conceptual contributions, analyses and interpretations remain the responsibility of the authors.